\pdfoutput=1

\documentclass[11pt]{article}

\usepackage[final]{acl}

\usepackage{times}
\usepackage{latexsym}
\usepackage{graphicx}

\usepackage[T1]{fontenc}

\usepackage[utf8]{inputenc}

\usepackage{microtype}

\usepackage{inconsolata}

\usepackage{graphicx}

\usepackage{booktabs}       
\usepackage{pifont} 
\usepackage{amsmath}
\usepackage{amsfonts}       
\usepackage{nicefrac}       
\usepackage{microtype}      
\usepackage{xcolor}         
\usepackage{adjustbox}
\usepackage{fontawesome}
\usepackage{float}
\usepackage{enumitem}
\usepackage{makecell}
\usepackage{comment}
\usepackage{array}
\usepackage{graphicx}
\usepackage{multirow}   
\usepackage{array}
\usepackage{tabularx} 
\usepackage{booktabs} 
\usepackage{subcaption}
\usepackage{hyperref}
\usepackage{multirow}
\usepackage{xspace}
\usepackage{soul}
\usepackage{placeins}
\usepackage{dblfloatfix}
\usepackage{cleveref}
\usepackage{xurl}

\newcommand{\CHRONOBERG}{Chronoberg\xspace}

\definecolor{highgreen}{RGB}{198,239,206}
\definecolor{medyellow}{RGB}{255,235,156}
\definecolor{lowred}{RGB}{255,199,206}
\definecolor{naGray}{RGB}{235,235,235}
\definecolor{sectionblue}{RGB}{68,114,196}

\definecolor{darkblue}{RGB}{46,25,110}
\newcommand{\dssectionheader}[1]{%
   \noindent\framebox[\columnwidth]{%
      {\fontfamily{phv}\selectfont \textbf{\textcolor{darkblue}{#1}}}
   }
}

\newcommand{\dsquestion}[1]{%
    {\noindent \fontfamily{phv}\selectfont \textcolor{darkblue}{\textbf{#1}}}
}

\newcommand{\dsquestionex}[2]{%
    {\noindent \fontfamily{phv}\selectfont \textcolor{darkblue}{\textbf{#1} #2}}
}

\newcommand{\dsanswer}[1]{%
   {\noindent #1 \medskip}
}

\title{\CHRONOBERG: A 250-Year Diachronic Corpus with Temporal Affective Lexicons for Language Model Analysis}

\author{
  \textbf{Niharika Hegde\textsuperscript{1,4*}},
  \textbf{Subarnaduti Paul\textsuperscript{2*}},
  \textbf{Lars-Joel Frey\textsuperscript{4}},
  \textbf{Manuel Brack\textsuperscript{1,3\ddag}},
\\
  \textbf{Kristian Kersting\textsuperscript{1,3,4,5}},
  \textbf{Martin Mundt\textsuperscript{2\dag}},
  \textbf{Patrick Schramowski\textsuperscript{1,3,4,6\dag}}
\\
\\
  \textsuperscript{1}DFKI,
  \textsuperscript{2}University of Bremen,
  \textsuperscript{3}hessian.AI,
\\
  \textsuperscript{4}TU Darmstadt,
  \textsuperscript{5}Centre for Cognitive Science, TU Darmstadt,
  \textsuperscript{6}CERTAIN
\\
  \small{
    \textbf{Correspondence:} \href{mailto:niharika.hegde@dfki.de}{niharika.hegde@dfki.de},\href{mailto:spaul@uni-bremen.de}{spaul@uni-bremen.de}
  }
}

\begin{document}
\maketitle
\begingroup
\renewcommand\thefootnote{}\footnotetext{%
\textsuperscript{*}Equal contribution.
\textsuperscript{\dag}Equal supervision.
\textsuperscript{\ddag}Work done while at DFKI and TU Darmstadt.}
\endgroup

\begin{abstract}
Large language models (LLMs) excel at operating at scale by leveraging social media and data crawled from the web. Whereas existing corpora are diverse, their frequent lack of long-term temporal structure may however limit an LLM's ability to contextualize semantic and normative evolution of language and to capture diachronic variation. To support the latter, we introduce \CHRONOBERG\footnote{\CHRONOBERG is publicly available on HuggingFace at \url{https://huggingface.co/datasets/spaul25/Chronoberg}, with code at \url{https://github.com/paulsubarna/Chronoberg}.}, a temporally structured corpus of English book texts spanning 250 years, curated from Project Gutenberg and enriched with a variety of temporal annotations. First, the edited nature of books enables us to quantify lexical semantic change through time-sensitive Valence-Arousal-Dominance analysis and to support historically-grounded interpretation. With the lexicons at hand, we demonstrate a need for modern LLM-based tools to better situate their contextualization of sentiment and detection of discriminatory language across various time-periods. In fact, we show how these tools struggle with diachronic shifts in meaning, emphasizing the need for temporally aware training and evaluation pipelines, and positioning \CHRONOBERG as a scalable resource for respective studies.
\end{abstract} 

\noindent
\textcolor{red}{Disclaimer:} This paper includes language and display of samples that could be offensive to readers.

\section{Introduction}
Language evolves continuously, reflecting shifts in knowledge, culture, and social norms. Yet, most large language models (LLMs) are trained on near-stationary datasets. Existing web-crawled corpora such as Common Crawl and Wikipedia best feature short-horizon temporal variations. Alas, without temporal grounding, language models risk conflating historical and contemporary meanings, for example, misinterpreting phrases such as ``\textit{Where is the woman to strew the flowers?}'' by applying modern connotations. Such misreadings can distort semantic understanding, but can also amplify outdated stereotypes and ethical blind spots \citep{blodgett2020survey}. As language and societal norms continue to evolve, it becomes increasingly critical to understand how models adapt their representations to ongoing and possible future linguistic change \citep{dhingra2022time}. 

This challenge is complex, as these shifts operate over centuries, not months. Semantic change ranges from evaluative shifts in affective polarity (``pejoration'', ``amelioration'') to non-evaluative shifts in meaning scope (``metaphor'', ``narrowing'', ``generalization'') \citep{jeffers1979principles,traugott2017semantic}. Existing resources \citep{googlebooks, davies2008coca, davies2010coha, jang2022temporalwiki} do not simultaneously provide the scale, temporal depth, and structured annotation needed to study them, or analyse where modern classifiers fail because of them. 
For such temporal contextualization, collections of edited books as a form of curated archive provide a more suitable resource, as demonstrated by pioneering ``culturomics'' studies \citep{michel2011google}. However, no existing book-derived resources simultaneously satisfy the above requirements, while also being openly accessible. Large n-gram archives  \citep{michel2011google,eebo} discard sentence context; curated corpora \citep{davies2010coha,alatrash2020ccoha} are too small or remain closed-source; and recent temporal datasets \citep{li2025tic,jang2022temporalwiki} primarily target short-horizon factual drift.

To bridge this gap, we introduce \CHRONOBERG, a diachronic corpus of 2.7 billion tokens spanning 250 years of English literary texts (1750--2000), curated from Project Gutenberg \citep{pgh} and enriched with temporal annotations. Three contributions are central. First, we construct temporally calibrated Valence-Arousal-Dominance (VAD) lexicons covering over 300,000 words across five 50-year epochs by propagating human-annotated VAD scores \citep{mohammad2018vad}. Together with sentence-level VAD annotations for the full corpus, these lexicons enable fine-grained tracking of sentiment trends across time. Second, we validate the lexicons against words with documented evaluative and non-evaluative changes from the linguistics literature \citep{jeffers1979principles,hamilton2016diachronic,schlechtweg2020semeval}, showing that evaluative shifts with clear VAD signal are correctly detected. Third, we demonstrate a practical consequence of temporal misalignment: contemporary LLM-based hate-speech classifiers \citep{liu2019roberta, lees2022perspectiveapi} conflate modern connotations with historical reality and how contemporary language models perform under temporal shift. \CHRONOBERG, including all annotations, will be made publicly available, supporting future work in diachronic lexical analysis and temporally aware machine learning and evaluation. 
\section{\CHRONOBERG Dataset} 
\begin{figure*}[t!]
    \centering
    \includegraphics[width=\linewidth]{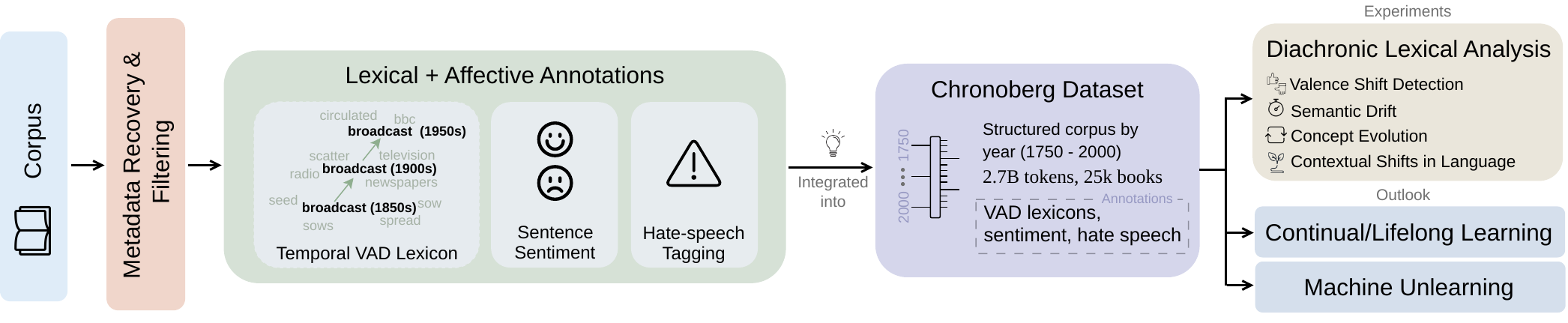}
    \caption{Overview of the \CHRONOBERG dataset pipeline, spanning corpus curation, metadata recovery, and diachronic lexical analysis. The resulting annotations, including VAD lexicons and sentiment scores, form an integral part of the dataset and support downstream machine learning investigation.}
    \label{fig:cb_pipeline}
\end{figure*}

We describe the construction of \CHRONOBERG, a temporal corpus designed to study the semantic evolution of the English language across centuries. We first situate it within existing resources, before detailing the pipeline, depicted in \Cref{fig:cb_pipeline}, covering curation, metadata inference, preprocessing, filtering, and VAD lexicon creation. 

\label{sec:dataset}

\subsection{Relation to Existing Works}
Existing temporal corpora fall short in different ways. Large n-gram archives such as Google Books and Early English Books (EEBO) \citep{michel2011google, eebo, eebo_tcp} offer broad historical coverage but discard sentence-level context, whereas curated corpora such as COHA \citep{davies2008coca}, COCA \citep{davies2010coha} and CCOHA \citep{alatrash2020ccoha} provide richer annotations but are not freely accessible at scale. Similarly, more recent works in temporal dataset construction \citep{li2025tic, jang2022temporalwiki} only target short-horizon factual drift, leaving long-horizon connotative evolution largely understudied computationally, despite a rich, extensive theoretical grounding in historical linguistics.

On the theoretical side, historical linguistics has established a principled typology of meaning change \citep{bloomfield1933language,jeffers1979principles,lehrer1985markedness,traugott2001regularity}, documented in resources such as the Historical Thesaurus of the OED \citep{kay2009historicaloed}. Computationally, distributional methods detect semantic shift at scale \citep{gulordava2011distributional,jatowt2014framework,kulkarni2015statistically}, culminating in word embedding approaches that reveal statistical laws of change \citep{hamilton2016diachronic} and transformer models that resolve individual senses over time \citep{schlechtweg2020semeval,edmondson2022word,cassotti2024using}.

These methods characterise whether and how meanings shift, but not the affective direction of that shift. This affective dimension is captured by Valence-Arousal-Dominance (VAD) frameworks, which encode polarity, activation level, and perceived control \citep{osgood1957,russell2003core}, extended to tens of thousands of words through human annotation \citep{mohammad2018vad, mohammad2025nrcvadlexiconv2}. These lexicons, however, are synchronic by construction, reflecting contemporary usage only.

This is consequential because, without connecting semantic shift to affective direction, downstream tasks such as hate-speech detection, sentiment analysis, and social bias auditing cannot account for the fact that the same word may carry different charge in a different era. This limitation is compounded by evidence that shifting social attitudes become embedded in NLP systems \citep{dehumanization,Queerinai2023} and that rapidly evolving language undermines classifier robustness \citep{keidar2022slangvolution,mehta2025genalpha}. \CHRONOBERG addresses these gaps by combining full-text coverage, yearly temporal granularity, long-horizon historical depth, and structured affective annotation over a 250-year period.

\subsection{Collection Methodology}
We build \CHRONOBERG on \citep{pgh}, an openly accessible corpus of literary texts that provides copyright-free English works at scale, in plain text and HTML formats, with metadata accessible via an API. 
However, the metadata is not directly suitable for temporal analysis. Original publication dates are frequently absent, incomplete, or inaccurate; available dates often reflect digitization, cataloguing, or release events rather than first publication. This poses a significant challenge for diachronic analysis, since assigning texts to incorrect years would directly distort estimates of semantic and affective change. \\

\noindent \textbf{Metadata Recovery.}
We developed an inference pipeline that combines internal metadata with queries to external bibliographic sources (\citet{openlibrary} and \citet{wikipedia}) and checks the consistency of inferred years against authors' lifespans. Manual verification of 100 randomly sampled books (spanning years 1611-1912) identified OpenLibrary as the most reliable source, achieving a mean absolute error (MAE) of $\pm3.05$ years and standard deviation (SD) of $5.20$ years (see Appendix~\ref{sec:datacuration}). Although this yields an uncertainty of 3-5 years, the margin is acceptable since our analysis operates at the scale of decades rather than years, comfortably exceeding the variance. We accordingly recommend that future studies adopt temporal bins no smaller than 15 years. \\

\noindent \textbf{Filtering.}
Using OpenLibrary as the default inference source based on the prior validation results, we excluded books without an inferrable publication year or known author, discarded posthumous editions via author lifespan checks, and retained only English-language texts published between 1750 and 2000. The resulting chronological backbone of \CHRONOBERG contains 25,061 out of 73,500 books in Project Gutenberg, allowing year-by-year aggregation, annotation, and subsequent analysis.

\subsection{Lexical and Affective Annotations}\label{sec:vad_construction}

We next describe the construction of \CHRONOBERG's core analytical contribution: a set of temporally aligned lexicons that tracks semantic change in word meanings across the 250 years. Methodologies for studying semantic change have evolved from qualitative, manual linguistic analysis \citep{bureal, Ullman} toward large-scale distributional approaches, with alignment techniques such as PPMI \citep{bullinaria2007semantic}, Word2Vec \citep{mikolov2013word2vec}, SVD \citep{levy2015svd} and CADE \citep{bianchi2020cade} enabling reliable recovery of directional semantic shifts that experts can verify \citep{blank1997, kulkarni2015statistically, hamilton2016diachronic, schlechtweg2020semeval, cassotti2024using}. These methods, however, only characterise whether and how meanings shift, but not the affective direction of that shift.

Influential factor analysis \citep{osgood1957, russell2003core} addresses this by decomposing affective meaning into three interpretable dimensions: Valence, Arousal, and Dominance, with human-annotated VAD lexicons extending this framework to 45,000 contemporary English words \citep{mohammad2018vad, mohammad2025nrcvadlexiconv2}. These lexicons, however, are synchronic by construction and cannot track the shifting connotations of words (e.g., \textit{broadcast} or \textit{febrile}) whose historical usage modern ratings systematically misrepresent \citep{perc2012evolution}. In \CHRONOBERG, we address this by combining diachronic semantics with human-annotated VAD scoring, proposing a novel set of temporally aligned lexicons spanning over 300,000 words across five 50-year epochs alongside sentence-level VAD scores for the full corpus.\\

\noindent \textbf{Temporal VAD Lexicons.} 
We model the shifts in word meaning using diachronic distributional semantics, learning a high-dimensional vector for each word from its co-occurrence patterns within a given time period.

Concretely, we train separate Word2Vec \citep{mikolov2013word2vec} models on a temporal slice of the \CHRONOBERG corpus, with each slice corresponding to a 50-year interval between 1750 and 2000, chosen according to the observed variance in publication data estimation and including a reasonable safety buffer. Following validation to select suitable hyperparameters, training for each interval has been conducted for 10 epochs with a context window size of 5 tokens on either side and an embedding vector dimensionality of 300. We note that interval  boundaries do not correspond to any specific historical eras or expert-defined periodization, and leave these to future work. The resulting embedding spaces are subsequently aligned using Compass Aligned Distributional Embeddings (CADE) \citep{bianchi2020cade}, which facilitates the direct comparison of word vectors across different decades by anchoring them to a shared compass model of the full corpus.

We chose Word2Vec + CADE for being computationally efficient over 250 years of data and reliable for diachronic nearest-neighbour retrieval. This aligns with our VAD propagation method, which requires stable vector-space alignment to preserve local geometric consistency for cross-temporal semantic neighbourhood retrieval, a property that CADE's compass-based alignment is specifically designed to ensure. While transformer embeddings \citep{bert} could offer richer contextual representations, we do not expect fundamental shifts in VAD interpretation, as the core compression principles remain unchanged.

We then estimate VAD scores for each target ($w$) by selecting the Top-K nearest neighbours ($\mathcal{N}_k(w)$) in the embedding space and averaging their corresponding VAD values from the human-annotated NRC VAD lexicon \citep{mohammad2018vad, mohammad2025nrcvadlexiconv2}:
\begin{equation}
\begin{aligned}
\mathcal{N}_k(w)
  &= \operatorname{Top}\!k_{\,u\in \mathcal{V}_{\text{VAD}}\setminus\{w\}}
     \; s(e_w,e_u), \\
\widehat{\mathbf{A}}_{\text{VAD}}(w)
  &= \frac{1}{|\mathcal{N}_k(w)|}
     \sum_{u\in \mathcal{N}_k(w)} \mathbf{A}_{\text{VAD}}(u).
\end{aligned}
\end{equation}
Here, $\mathbf{e}_w \in \mathbb{R}^d$ is the embedding of target word $w$,
$s(e_w,e_u)$ is the cosine similarity, and $\mathcal{V}_{\text{VAD}}$ the set of words with NRC VAD annotations $\mathbf{A}_{\text{VAD}}(u)\in\mathbb{R}^3$; the target word itself is excluded from its own neighbourhood set during propagation. We thus make use of the human-annotated scores in the NRC VAD lexicon, which primarily reflect contemporary interpretations of words, but re-contextualize them computationally to account for historical contexts or diachronic semantic shifts that certain words may have undergone. To this end, we use modern valence scores as anchors to detect relative semantic drift. While this may introduce systematic anachronisms, it is unavoidable and presents a natural choice, as retrospective human annotations at scale from several hundreds of years ago are practically unattainable, and few, if any, historians can operate at such a scale. For our purposes, temporal VAD lexicons thus serve as a measure of directional valence trajectories, rather than an absolute historical ground truth.  

We thereby present temporally adjusted lexicons spanning 335,804 words, each assigned a real-valued score between -1 and 1 for the three VAD dimensions. A key challenge in this process lies in determining a suitable number of top-K neighbours. Selecting too few (top-10) can miss contextual diversity, but taking into account too many (top-500) may lead to semantic over-smoothing and can introduce noise. Following empirical analysis, averaging the scores from the top-20 neighbours seems to mitigate these adverse effects. To determine the number of neighbours required to retrieve 20 known words from the lexicon, we analysed the retrieval rates for 100 anchor words of high and low valence from \CHRONOBERG. This analysis further revealed that retrieving 20 known neighbours typically necessitates inspecting the top-100 nearest neighbours (see Appendix \ref{app:section_neighbors}).\\

\begin{figure}[t!]
    \centering
    \includegraphics[width=\columnwidth]{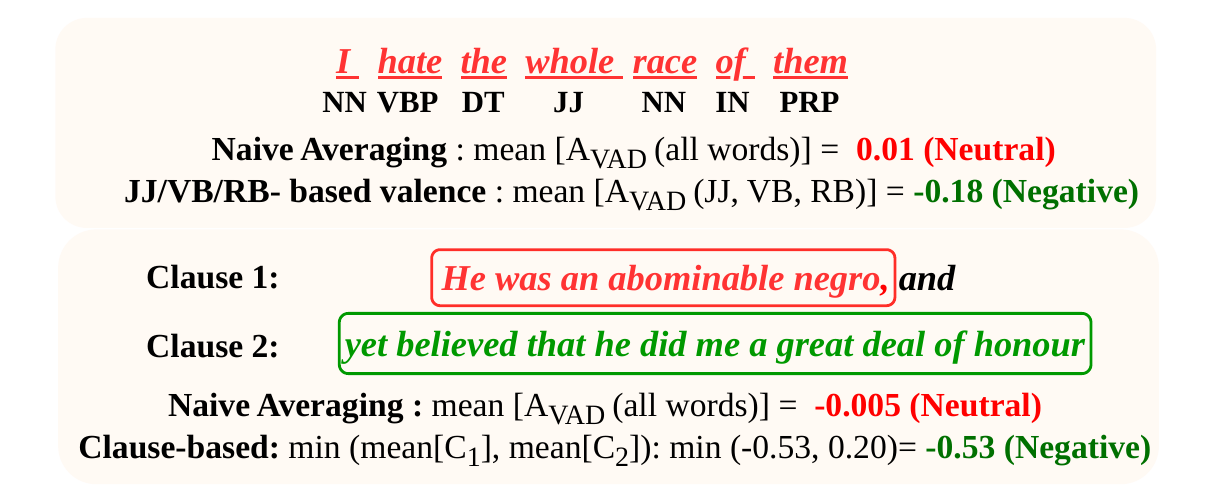}
    \caption{Sentence-level VAD scoring in \CHRONOBERG. Instead of simple averaging across all words, we introduce a two-stage aggregation strategy: (i) part-of-speech--based
    averaging over adjectives, verbs, and adverbs (JJ/VB/RB), followed by (ii) clause-based scoring, which computes clause-level valence means and selects the most extreme
    value.}
    \label{fig:sentence_level_VAD}
\end{figure}
\noindent \textbf{Sentence-level VAD Annotations.} Extending our analysis beyond individual words, we leverage our five temporal VAD lexicons to assess sentiment in full sentences. However, a naive aggregation of word scores can lead to a neutrality bias, given the high frequency of neutrally perceived words in language. To mitigate this issue, we introduce two key elements in our scoring pipeline: JJ/VB/RB and clause-based averaging, illustrated in Figure \ref{fig:sentence_level_VAD}.

\textit{JJ/VB/RB averaging: } First, we focus only on emotionally salient parts of speech (in particular verbs (VB), adverbs (RB), and adjectives (JJ)), thereby minimizing the influence of neutral words on the final sentence score. 

\textit{Clause-based averaging: } Second, our method accounts for sentiment variations within a sentence, following prior work \citep{clause-avg}. Accordingly, we calculate an average valence for each clause, based on its adjectives (JJ), verbs (VB), and adverbs (RB). The overall sentence score is then the minimum value among these clause-level scores.

\textit{Final Score:} We compute the final score $\widehat{\mathbf{l}}_{\text{VAD}}(\text{sent})$ by combining these two approaches:
\begin{equation}
\min_{C_i \in \text{Clauses}}
\left( 
\frac{\sum_{t \in C_i, \ \text{pos}(t) \in \{\text{JJ,VB,RB}\}} \mathbf{A}_{\text{VAD}}(t)}
{N_{C_i,\{\text{JJ,VB,RB}\}}}
\right),
\label{eq: valence_scores}
\end{equation}
where $\mathbf{A}_{\text{VAD}}(t)$ represents the VAD scores and pos(t) is the part of speech tag for each token in the sentence. 
Since values range from -1 to +1, we will consider the sign to carry a respective connotation for simplicity in further analysis.  This approach is particularly effective on sentences that combine positive and negative clauses, where naive averaging would otherwise mask the hateful content. We believe this to be justified by focusing on the assessment of the overall change in consecutive analysis. However, we acknowledge that perceived connotation can depend on various subjective factors. Our sentence-level valence annotations are a core component of \CHRONOBERG and are publicly available to support transparency and further analysis.
%

\subsection{\CHRONOBERG Composition \& Statistics} 
Finally, we highlight \CHRONOBERG's summary statistics. A dataset sheet for datasets \citep{gebru2021datasheet} is provided in Appendix \ref{sec:datasheet_for_datasets}. \\

\noindent \textbf{Composition.}
Overall, \CHRONOBERG is composed of 2.7B tokens, representing 91M sentences from 25,061 English-language books published between 1750 and 2000, with additional metadata in the form of temporally-aligned VAD lexicons that span 335,804 words. 
On average, approximately 28\% of sentences per 50-year epoch are classified as \textit{negative} (valence $< 0 - \epsilon$), while 50\% are \textit{positive} (valence $> 0 + \epsilon$), considering $\epsilon$ to be 0.05. Notably, some unique samples across epochs exhibit a change in average valence scores. \\

\noindent \textbf{Diachronic Shifts in Words \& Sentences.}
\label{subsec:stats}
To distinguish semantically stable words from those undergoing genuine diachronic drift, we apply a threshold $|\Delta|>0.1$ to the endpoint difference between first and last available epoch valence estimates. A word is classified as \textit{Shifting} if $|v_\text{last} - v_\text{first}| > 0.1$ and the valence category changes (positive $\leftrightarrow$ negative $\leftrightarrow$ neutral); otherwise it is classified as \textit{Stable}. For the validation study in \ref{sec:validation}, we further assign a confidence tier to each evaluative word based on two thresholds applied jointly to local step-wise movement and net endpoint shifts: (i) \colorbox{highgreen}{\textbf{H}}igh requires at least two consecutive deltas exceeding the noise floor \textit{and} $|\Delta_\text{net}| > 0.2$: the VAD trajectory shows a clear, multi-step directional shift. (ii) \colorbox{medyellow}{\textbf{M}}edium is assigned when local movement is present
but the net shift is modest ($|\Delta_\text{net}| \leq 0.2$), or when only a single significant step is observed: the VAD signal is detectable but not unambiguous. (iii) \colorbox{lowred}{\textbf{L}}ow covers two cases: words where no individual step exceeds the noise floor but sub-threshold drift accumulates directionally across epochs. Words with no significant movement on either measure are classified \textit{Stable}. Across the full \CHRONOBERG lexicon of 335,804 words, 44,049 are assigned \colorbox{highgreen}{\textbf{H}}igh, 36,301 \colorbox{medyellow}{\textbf{M}}edium, and 2,518 \colorbox{lowred}{\textbf{L}}ow, with the remainder classified Stable, indicating that roughly one quarter of the lexicon shows detectable affective drift over the 250-year window.
\begin{table}[t!]
\centering
\small
\setlength{\tabcolsep}{1.7pt}
\renewcommand{\arraystretch}{0.85}
\begin{tabular}{l l c r r r r r}
\toprule
\textbf{Shift} & \textbf{Word} & \textbf{Tier}
& \textbf{1750s} & \textbf{1800s} & \textbf{1850s}
& \textbf{1900s} & \textbf{1950s} \\
\midrule
\multirow{5}{*}{\rotatebox[origin=c]{90}{\emph{pos$\to$neg}}}
 & asylum      & \colorbox{highgreen}{\textbf{H}} &  0.27 & -0.24 & -0.54 & -0.52 & -0.65 \\
 & weird       & \colorbox{highgreen}{\textbf{H}} &  0.30 &  0.01 & -0.28 & -0.33 & -0.43 \\
 & homeless    & \colorbox{medyellow}{\textbf{M}} &  0.11 & -0.62 & -0.63 & -0.66 & -0.56 \\
 & punk        & \colorbox{medyellow}{\textbf{M}} &  0.20 &  0.14 & -0.25 & -0.17 & -0.26 \\
 & erasure  & \colorbox{lowred}{\textbf{L}} &  0.10 &  0.02 & -0.06 & -0.12 & -0.13 \\
 & astrologer  & \colorbox{lowred}{\textbf{L}} &  0.30 &  0.20 & 0.11 & 0.04 & 0.00 \\
\midrule
\multirow{5}{*}{\rotatebox[origin=c]{90}{\emph{neg$\to$pos}}}
 & febrile      & \colorbox{highgreen}{\textbf{H}} & -0.58 & -0.53 & -0.66 & -0.54 &  0.33 \\
 & infatuation  & \colorbox{highgreen}{\textbf{H}} & -0.66 & -0.63 & -0.52 & -0.35 &  0.40 \\
 & shack & \colorbox{medyellow}{\textbf{M}} & -0.67 & -0.05 & -0.02 & 0.30 &  0.10 \\
 & managerial & \colorbox{medyellow}{\textbf{M}} & -0.65 & 0.25 & 0.20 & 0.26 &  0.30 \\
 & heterogeneous  & \colorbox{lowred}{\textbf{L}} &  -0.21 &  -0.17 & -0.13 & -0.03 & 0.07 \\
 & modernism  & \colorbox{lowred}{\textbf{L}} &  0.08 & 0.14 & 0.20 & 0.20 & 0.30 \\
\bottomrule
\end{tabular}
\caption{Examples of words with strong diachronic valence shifts in \CHRONOBERG. Scores are averaged over top-20 nearest neighbours per epoch and range from $-1$ (maximally negative) to $+1$ (maximally positive). Confidence tiers: \colorbox{highgreen}{\textbf{H}} = unambiguous, dominant VAD shift; \colorbox{medyellow}{\textbf{M}} = partial/single-step shift; \colorbox{lowred}{\textbf{L}} = sub-threshold drift.}
\label{table:shifts}
\end{table}

Table~\ref{table:shifts} presents representative examples illustrating all three tiers. Among \textit{pos$\to$neg} words, \textit{asylum} ($+0.27{\to}-0.65$, \colorbox{highgreen}{\textbf{H}}) pejorated as the word shifted from sanctuary to associations with mental institutions; \textit{homeless} ($+0.11{\to}-0.56$, \colorbox{medyellow}{\textbf{M}}) acquired negative valence through urbanisation, concentrated in a single epoch; \textit{astrologer} ($+0.30{\to}+0.00$, \colorbox{lowred}{\textbf{L}}) shows slow sub-threshold drift. Among \textit{neg$\to$pos} words, \textit{febrile} ($-0.58{\to}+0.33$, \colorbox{highgreen}{\textbf{H}}) ameliorated as metaphorical usage overtook clinical sense; \textit{managerial} ($-0.65{\to}+0.30$, \colorbox{medyellow}{\textbf{M}}) gained prestige; and \textit{heterogeneous} ($-0.21{\to}+0.07$, \colorbox{lowred}{\textbf{L}}) drifted slowly toward neutral as diversity became descriptive rather than evaluative.

These examples emphasize the value of our temporal VAD lexicons and \CHRONOBERG as instruments for fine-grained analysis of semantic and affective change. Additional examples are in Appendix \ref{app:dia_examples}.

\section{Experiments and Dataset Analysis}
\label{sec:experiments}
We present three experiments demonstrating \CHRONOBERG's utility: validating the temporal VAD lexicons against words with documented semantic shifts, to assess if VAD trajectories capture such shifts; examining whether contemporary hate-speech detectors capture shifts in harmful language as seen in \CHRONOBERG; and investigating the extent to which these systems rely on surface-level lexical cues rather than contextual understanding.

\subsection{External Validation of the VAD Lexicons}
\label{sec:validation}
To assess whether \CHRONOBERG's temporal VAD scores capture documented shifts, we validate against 119 words with verified changes compiled from existing literature, covering both evaluative cases (pejoration, amelioration, contronym) and non-evaluative cases (metaphor, narrowing, generalization, subjectification, polysemy). Words whose shift falls outside \CHRONOBERG's measurement window are excluded as outliers ($n{=}12$). Each word is assigned a confidence tier (\colorbox{highgreen}{\textbf{H}}/\colorbox{medyellow}{\textbf{M}}/\colorbox{lowred}{\textbf{L}}) using the criteria defined in \ref{subsec:stats}, and each misrepresentation is assigned to one of three error categories: \emph{direction mismatch} ($\ddagger$) where the corpus-level trajectory runs counter to the documented direction of change, typically because a dominant literary meaning masks a shift that unfolded in spoken or informal registers; \emph{register gap} ($\star$), where the shifted meaning is under-represented in edited book text to produce measurable valence shift; and \emph{bidirectional sense split} ($\dagger$), where opposing senses coexist and cancel in the aggregate score.

\begin{table}[t!]
\centering
\small
\setlength{\tabcolsep}{4pt}
\textbf{(a) Evaluative changes}\\[4pt]
\begin{tabular*}{\columnwidth}{@{\extracolsep{\fill}} l l r r l}
\toprule
\textbf{Type} & \textbf{Conf.\ Tier}
  & \textbf{Rep.} & \textbf{Misrep.} & \textbf{Reason} \\
\midrule
\multirow{3}{*}{Pejoration (30)}
  & \colorbox{highgreen}{H}  &  8 &  1 & $\ddagger$ \\
  & \colorbox{medyellow}{M}  &  7 &  9 & $\ddagger$ \\
  & \colorbox{lowred}{L}             &  0 &  5 & $\star$    \\
\midrule
\multirow{2}{*}{Amelioration (9)}
  & \colorbox{highgreen}{H}  &  4 &  1 & $\ddagger$ \\
  & \colorbox{medyellow}{M}  &  4 &  0 & --         \\
\midrule
Contronym (1)
  & \colorbox{medyellow}{M}  &  0 &  1 & $\dagger$  \\
\midrule
\textbf{Total (40)} & & \textbf{23} & \textbf{17} & \\
\bottomrule
\end{tabular*}

\vspace{6pt}
\textbf{(b) Non-evaluative changes}\\[4pt]
\begin{tabular*}{\columnwidth}{@{\extracolsep{\fill}} l r r}
\toprule
\textbf{Type ($n$)}
  & \textbf{Val.\ stable}
  & \textbf{Val.\ shifts} \\
\midrule
Metaphor (21)                  &  5 & 16 \\
Narrowing (4)                  &  0 &  4 \\
Generalization (3)             &  1 &  2 \\
Subjectification (7)           &  1 &  6 \\
\midrule
\multicolumn{3}{l}{\textit{Multiple coexisting senses}} \\
\quad Polysemy (5)             &  0 &  5 \\
\quad Domain / Technology (3)  &  1 &  2 \\
\midrule
SemEval Stable (24)            &  9 & 15 \\
\midrule
\textbf{Total (67)} & \textbf{17} & \textbf{50} \\
\bottomrule
\end{tabular*}
\caption{Validation of \CHRONOBERG's thresholded VAD lexicons against documented semantic shifts. In \textbf{(a)} Confidence tiers (Conf.\ Tier): \colorbox{highgreen}{\textbf{H}}igh = unambiguous, dominant VAD shift in corpus; \colorbox{medyellow}{\textbf{M}}edium = partial or single-step shift; \colorbox{lowred}{\textbf{L}}ow = sub-threshold drift invisible to VAD. In \textbf{(b)}, \textit{Val.\ stable} = valence correctly flat, \textit{Val.\ shifts} = connotative drift driven by a shift in the distribution of coexisting word senses. \textbf{Note}: ($\ddagger$): direction mismatch: corpus-level trajectory runs counter to expected direction (e.g.\ \textit{bully}, \textit{guy}). ($\star$): register gap: shifted sense underrepresented in \CHRONOBERG\ (e.g.\ \textit{Fairy}, \textit{Bent}) or obscured by a stable dominant sense (e.g.\ \textit{Queen}). ($\dagger$): bidirectional sense split: opposing senses coexist and cancel (e.g.\ \textit{sanction}).}
\label{tab:vad_validation}
\end{table}

\begin{table*}
\centering
\resizebox{\linewidth}{!}{%
\begin{tabular}{>{\raggedright}p{1.cm}p{10.6cm}ccccp{1cm}c}
\toprule
\multirow{2}{*}{\textbf{YEAR}} & \multirow{2}{*}{\textbf{Sentences}} & \multicolumn{4}{c}{\textbf{Hate-Check Models}} & \textbf{Valence} & \textbf{Affective} \\
& & \textbf{RoBERTA+Persp} & \textbf{OpenAI} &\textbf{Qwen}  &\textbf{Mistral} & \textbf{Score} & \textbf{Sentiment} \\
\midrule

1750s & but I loathe you, you apache indian! & \textcolor{red}{\faFlag} & \textcolor{red}{\faFlag} & \textcolor{red}{\faFlag}  & \textcolor{red}{\faFlag}  &  -0.50 & \faThumbsODown \\
& Where is the woman to strew the flowers? & \textcolor{red}{\faFlag} & \color{green}\ding{51} &\color{green}\ding{51}  & \color{green}\ding{51}  & 0.14 & \faThumbsOUp\\
& you horse-hair hypocrite, you! & \color{green}\ding{51} & \color{green}\ding{51} & \textcolor{red}{\faFlag} & \color{green}\ding{51}  & -0.60 & \faThumbsODown \\

\midrule
1800s & How I wish that you were black!--I detest your colour. & \textcolor{red}{\faFlag} & \textcolor{red}{\faFlag} & \textcolor{red}{\faFlag} & \textcolor{red}{\faFlag} & -0.79 & \faThumbsODown \\
& The black parts of them must be cut away, when the skin is taken off. & \textcolor{red}{\faFlag} & \color{green}\ding{51} & \textcolor{red}{\faFlag} & \textcolor{red}{\faFlag} & 0.02 & \faThumbsOUp \\
& he redoubled his gayety and carelessness. & \color{green}\ding{51} & \color{green}\ding{51} & \color{green}\ding{51} & \color{green}\ding{51}  & -0.71 & \faThumbsODown \\

\midrule
1850s & The Irish are mostly very filthy and diseased. & \textcolor{red}{\faFlag} & \textcolor{red}{\faFlag} & \textcolor{red}{\faFlag} & \textcolor{red}{\faFlag} & -0.66 & \faThumbsODown \\
& They carry faggots. & \textcolor{red}{\faFlag} & \textcolor{red}{\faFlag} & \textcolor{red}{\faFlag} & \textcolor{red}{\faFlag} & 0.08 & \faThumbsOUp \\
& An outlandish pair brought from New York for some tory hussy & \color{green}\ding{51} & \color{green}\ding{51} & \color{green}\ding{51} & \textcolor{red}{\faFlag} & -0.10 & \faThumbsODown \\

\midrule
1900s & As a race, as a family, the blacks have no loyalty. & \textcolor{red}{\faFlag} & \textcolor{red}{\faFlag} & \textcolor{red}{\faFlag} & \textcolor{red}{\faFlag} & -0.02 & \faThumbsODown \\
& I may cut you out of my gold expedition, if you get gay. & \textcolor{red}{\faFlag} & \textcolor{red}{\faFlag} & \textcolor{red}{\faFlag} & \textcolor{red}{\faFlag} & 0.175 & \faThumbsOUp\\
& We know how heartless he is, how vindictive, how horribly cruel. & \color{green}\ding{51} & \color{green}\ding{51} & \textcolor{red}{\faFlag} & \color{green}\ding{51} & -0.77 & \faThumbsODown\\

\midrule
1950s & I hope with the Negroes, Indians, .. reduce Virginia.. & \textcolor{red}{\faFlag} & \textcolor{red}{\faFlag} & \textcolor{red}{\faFlag} & \textcolor{red}{\faFlag} & -0.18 & \faThumbsODown\\
& Black should never be worn at a wedding. & \textcolor{red}{\faFlag} & \textcolor{red}{\faFlag} & \color{green}\ding{51} & \color{green}\ding{51} & 0.10 & \faThumbsOUp \\
& Why are the Africans in Algeria rising against their white French oppressors? & \color{green}\ding{51} & \color{green}\ding{51} & \color{green}\ding{51} & \color{green}\ding{51} & -0.05 & \faThumbsODown \\

\bottomrule
\end{tabular}}
\caption{Sentence-level classifications in \CHRONOBERG using LLM-based hate-check tools (RoBERTa+Perspective API, OpenAI Moderation API, Qwen and Mistral) (\textcolor{red}{\faFlag} = Hate, \textcolor{green}{\ding{51}} = Non-hate) and valence-based scoring (\faThumbsODown = Negative, \faThumbsOUp = Positive sentiment). For each 50-year interval, we organize based on model agreement. The first row illustrates cases where all tools collectively classify an instance as harmful. Rows 2-3 show instances where they disagree, classifying them as either positive or negative. While VAD lexicons provide interpretable complementary signals, we acknowledge that harmful texts are inherently subjective; therefore, we do not regard them as definitive solutions to LLM misclassification, but rather as potential tools to enhance performance.}
\label{tab:hate_speech_condensed}
\vspace{0.5em}
\end{table*}
%

Table~\ref{tab:vad_validation}(a) shows that 23 of 40 in-scope \emph{evaluative} words are correctly represented, with agreement between observed valence trajectory and expected direction of change confirmed against expert annotations. Accuracy stratifies by tier: \colorbox{highgreen}{\textbf{H}}-tier words are correct in 12 of 14 cases (86\%), \colorbox{medyellow}{\textbf{M}}-tier in 11 of 21 (52\%), and \colorbox{lowred}{\textbf{L}}-tier in none, indicating that the lexicons are reliable when corpus signal is strong and that the weaker aggregate is driven by inherently ambiguous medium-confidence cases. Of the 9 \colorbox{highgreen}{\textbf{H}}-tier pejorations, 8 are correctly represented (e.g., \textit{awful}; \textit{bastard}). \colorbox{medyellow}{\textbf{M}}-tier captures an additional 7 pejorations (e.g., \textit{bitch, dyke, prick}) where signal is present but weak or concentrated in a single epoch transition. For amelioration, 8 of 9 words are correctly represented: 4 at \colorbox{highgreen}{\textbf{H}}-tier (e.g., \textit{awesome}, \textit{witch}) and 4 at \colorbox{medyellow}{\textbf{M}}-tier (e.g., \textit{minx}, \textit{terrific}).
The 17 misrepresentations include one complex case belonging to \textit{contronyms} (\textit{sanction}), whose bidirection split is understandably diluted in the VAD lexicons; the other 16 misrepresentations fall into two interpretable categories. 11 words show corpus-level movement in the \textit{wrong direction} ($\ddagger$): 9 medium-tier pejorations (e.g.~\textit{bully} and \textit{egregious}) appear to ameliorate in book text because their dominant literary senses pull aggregate neighbourhood valence upward, masking the pejorative shift that occurred primarily in spoken and informal registers; the ameliorated word \textit{guy} and the pejorated word \textit{batty} similarly move in the opposite direction to expectation. Five words (\textit{fatal}, \textit{weasel}, \textit{queen}, \textit{bent}, \textit{fairy}) are assigned \colorbox{lowred}{\textbf{L}}ow: their pejorated senses were too infrequent in edited book text to produce measurable valence movement, reflecting a register gap ($\star$).
 
Table~\ref{tab:vad_validation}(b) depicts the result for non-evaluative changes, where no valence shift is expected. 8 words are correctly classified as valence-stable: 5 metaphor words (\textit{attack}, \textit{bug}, \textit{edge}, \textit{land}, \textit{platform}) and 1 each from generalization (\textit{head}), subjectification (\textit{major}), and domain/technology (\textit{computer}). The remaining 35 words not belonging to SemEval-2020 \citep{schlechtweg2020semeval} that show valence movement are analytically informative rather than erroneous: non-evaluative changes such as technological metaphors (\textit{network}, \textit{film}) and subjectification (\textit{actually}, \textit{starting}) produce genuine connotative drift as a word's dominant associative context shifts, pulling aggregate neighbourhood valence with it. Among the 24 SemEval words, 9 are correctly classified as stable and 15 show apparent valence movement. We note that SemEval-2020 defines stability over a short two-timepoint window, while \CHRONOBERG spans 250 years; thus, some of these cases may reflect genuine long-horizon drift that falls outside SemEval's detection threshold, rather than spurious correlation. Where misclassification does occur, it is attributable to representation gaps and polysemy dilution, which are expected limitations of a corpus-driven approach on historical text. These results also address a potential circularity in the propagation method \ref{sec:vad_construction}, since VAD scores are propagated from contemporary NRC anchors, they could in principle reflect modern rather than historical associations. However, the reliable recovery of documented shifts precisely where corpus signal is strong (that is, \colorbox{highgreen}{\textbf{H}}-tier cases) offers indirect evidence that such circularity does not dominate the propagated scores.

\subsection{Do LLMs Capture Shifts in Harmful Language within \CHRONOBERG?}
\label{sec:LLMs}

Having validated that the lexicons capture documented semantic change, we next leverage our temporal VAD annotations to contextualize practical notions of harmful language over time. Hate speech represents particularly well-defined and inherently negatively connoted subsets, where temporal shifts in meaning are especially consequential. We emphasize that lexical valence is neither equivalent to nor a proxy for hate. A sentence may be hateful without strongly negative valence and vice versa. The propagated VAD scores serve throughout as interpretable diagnostic signals of temporal misalignment, not as ground truth for hate detection. Prior works have shown that hate-speech classifiers degrade under temporal shift, as token-level sensitivity causes models to flag words regardless of historical usage \citep{davidson2017hate}. Cross-period generalisation also remains poor even with temporal lexical features \citep{rottger2021temporaladaptation, mcgillivray2022timedependent}, and static benchmarks actively mask both issues \citep{dibonaventura2025hatevolution}. This allows us to investigate alignment between hate-check outputs and VAD scores, highlighting cases where our annotations capture meaningful affective polarity. More importantly, discrepancies in sentiment expose where modern classifiers fail to recognize historically situated expressions of hate or over-generalize from present-day keyword associations.\\
To identify suitability, we first evaluated nine hate-speech detection tools on HateCheck \citep{rottger2021hatecheck} (see Appendix~\ref{sec:hatespeech}). Based on this analysis, we adopt a two-stage pipeline: RoBERTa\footnote{RoBERTa is a pretrained language model; we use ``hate-check tools'' as a collective shorthand.} \citep{liu2019roberta} achieves the strongest recall (P$=0.963$, R$=0.973$) but is prone to false positives, while Perspective API \citep{lees2022perspectiveapi} offers near-perfect precision at low recall (P$=0.993$, R$=0.389$); we therefore use RoBERTa to maximize coverage and Perspective API at a threshold of $\geq0.7$ to filter false positives. We also evaluate the OpenAI moderation API\footnote{\url{https://platform.openai.com/docs/guides/moderation}} as a recent baseline.

In Table~\ref{tab:hate_speech_condensed}, we show representative examples where VAD annotations and current hate-check tools agree and diverge. They are illustrative of notable trends in how modern hate-check tools flag negative sentences. While the latter achieve consistent correct predictions in cases of explicit sentiment (first row of each time interval), they seem to struggle when sentiment is implied rather than directly stated. For instance, a neutral phrase such as \textit{``Black should never be worn at a wedding''} is incorrectly flagged as hateful by both hate-check tools, whereas the valence scores more accurately capture its neutral sentiment (0.10). Another illustrative case from the 1850s is the phrase \textit{``tory hussy''}, where LLMs misinterpret \textit{hussy} with its modern connotation, influencing the judgment of the sentence’s hatefulness. 
\begin{figure}[t!]
   \centering
    \includegraphics[width=\columnwidth]{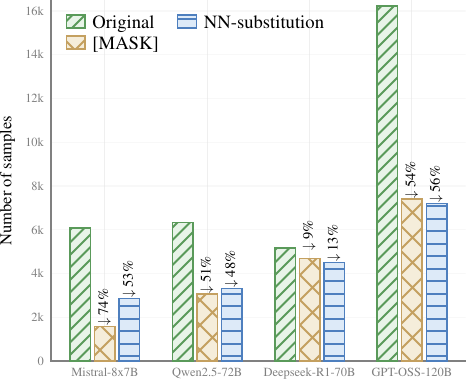}
    \caption{Substituting contemporary target words with historical equivalents improves the contextualization of modern LLMs for hate-speech identification. The green bars report the disagreement rates between LLMs and \CHRONOBERG's temporal VAD lexicons, while the orange and blue lines report the disagreement rates after applying two substitution strategies: replacing target words with a \texttt{[MASK]} token or a nearest-neighbour (NN) word from the corresponding era, across four models (Mistral, Qwen2.5, GPT, and DeepSeek). Both substitutions consistently reduce disagreement rates by 50-75\% for three of our four models, highlighting the models' over-reliance on surface-level lexical cues rather than historical context.}
    \label{fig:surface_cue}
\end{figure}

\subsection{Do Modern Hate-Check Tools Rely on Surface-Level Lexical Cues?}
Discrepancies between LLM hate-speech classifiers and our temporal VAD lexicons reveal a systematic pattern of surface-level token sensitivity. For instance, LLM classifiers negatively connotate contexts such as \textit{``They carry faggots''}, even though the term \textit{faggot} commonly referred to bundles of wood and not the contemporary association with slurs \citep{michel2011google}. In contrast, our temporal VAD lexicons assign the sentence a positive valence score, reflecting its historically situated meaning. This leads to a hypothesis that classifier decisions are driven by the target token itself, rather than its sentential context. To validate this hypothesis at scale, we conduct systematic substitutions of words with period-appropriate synonyms in the corresponding VAD lexicon. For instance, the word \textit{faggot} would be replaced with \textit{firewood}, \textit{sticks}, or \textit{bundle} in an otherwise identical sentence. Specifically, we first identify all sentences in \CHRONOBERG where RoBERTa and our VAD lexicons conflict, treating these disagreements as candidate cases of classifier overreach. We then screen these sentences across four contemporary LLMs  (Mistral \citep{jiang2023mistral7b}, Qwen2.5 \citep{qwen2.5}, GPT-OSS \citep{openai2025gptoss120bgptoss20bmodel}, Deepseek \citep{deepseekai2025deepseekr1incentivizingreasoningcapability}), applying a consistent substitution procedure: each flagged word is replaced with a historically accurate, semantically equivalent but non-toxic neighbor drawn from our Word2Vec embeddings,  under two conditions: replacing target words with a \texttt{[MASK]} token or a nearest-neighbor (NN) word from the corresponding era. In \Cref{fig:surface_cue}, we observe that a single substitution flips predictions from hateful to nonhateful for about $50-75\%$ of the initially flagged sentences for three of the four models. The results confirm that a simple lexical substitution is often sufficient to flip classifier predictions from hateful to not-hateful across most variants. Whereas the VAD score remains consistent throughout, modern classifiers remain sensitive to surface-level lexical cues and fail to account for diachronic shifts in word meaning.
\CHRONOBERG's temporal VAD lexicons can thus serve as a complementary diagnostic mechanism for identifying these failures. Beyond diagnosis, they point towards temporally aware adaptation: since continual pretraining on target-period text recovers within-period but not cross-period performance \citep{rottger2021temporaladaptation}, era-specific VAD trajectories could instead be provided to classifiers as an auxiliary signal indicating what a term meant and how it was perceived historically, directly targeting the surface-cue sensitivity demonstrated here. 
\section{Conclusion}
We introduced \CHRONOBERG, a large-scale, temporally structured corpus of English books spanning the years 1750-2000, enriched with diachronic VAD lexicons and sentence-level affective annotations. Using these resources, we quantified semantic and affective shifts, and demonstrated the need for modern LLM-based tools to better situate their detection of discriminatory language and contextualization of sentiment across various time periods. \CHRONOBERG opens several promising avenues, such as broader continual- and lifelong-learning studies with temporal training pipelines, as well as machine unlearning protocols to address historically contingent slurs or outdated facts. Future work includes strategic sampling of books containing homonyms or underrepresented words to update their meanings, exploring finer temporal granularities and drawing interdisciplinary connections to key historical eras or literary epochs.

\section{Limitations}
Although publication date inference introduces residual uncertainty (MAE $\pm3.05$, SD $5.20$ years), this is acceptable at the decadal granularity at which our analysis operates. We nonetheless recommend temporal bins no smaller than 15 years for future work. As retrospective human annotation at this scale is improbable, and expert linguistic validation across 335,804 words and 250 years equally impractical, the temporal VAD lexicons should be read as directional trajectory measures rather than absolute historical ground truth, with confidence tiers treated as indicative rather than definitive. Since \CHRONOBERG's literary corpus excludes the spoken and subcultural language where most pejorative shifts originate, failures on words such as \textit{fairy} and \textit{queen} are expected. Finally, although our substitution experiment demonstrates systematic surface-level sensitivity in LLM classifiers, hate speech is inherently subjective across historical periods and the binary hateful/non-hateful framing remains too coarse for nuanced historical analysis. Thus, the lexicons should accordingly be treated as a diagnostic tool in conjunction with moderation tools rather than as a replacement. 

\section{Ethical Considerations}
All texts in \CHRONOBERG are sourced from ~\citep{pgh}'s public domain collection, including some that reference identifiable historical figures and describe subpopulations by race, gender and religion; we caution against using the dataset to single out individuals or groups. The dataset contains historically situated language that readers may find offensive, including slurs and discriminatory expressions reproduced solely for scientific analysis. As the corpus reflects the literary landscape from the perspective of prolific authors between 1750 and 2000, findings should not be generalized beyond this time frame. The hate-speech annotations included in \CHRONOBERG are a diagnostic resource; they should not be used as training data for modern classifiers without careful mitigation, as doing so risks encoding historical prejudices into them. 
 
\bibliography{acl}
    
\clearpage
\appendix
\section*{Appendix}

We have organized our supplementary material in the following order:
\begin{itemize}
    \item \hyperref[sec:dataset_details_app]{\textbf{Section A: Dataset Curation and Lexical Analysis.}} Starting with an extended related works section, we then provide additional details on dataset collection, curation, and filtering pipeline, along with an extended analysis of VAD lexicons. 
    \item \hyperref[sec:hatespeech]{\textbf{Section B: Extended Hate-speech Analysis.}} Expanded evaluation of harmful texts in \CHRONOBERG, including comparisons across multiple hate-speech detection tools, sentence-level analysis, and highlighting cases of disagreement between different models and valence scores in hate prediction. 
    \item \hyperref[sec:datasheet_for_datasets]{\textbf{Section C: Datasheet for \CHRONOBERG.}}
\end{itemize}

\section{Additional \CHRONOBERG dataset details}
\label{sec:dataset_details_app}
This appendix complements main body Section \ref{sec:dataset} by providing additional information on the choice of data sources, available metadata, and recovery of publication dates through inference methods. Each of the following subsections expands upon specific components of the dataset pipeline to clarify design decisions and practical challenges. We also provide insights into the distribution and the thematic composition of \CHRONOBERG.

\subsection{Extended Related Works}
\label{sec:related_work_extended}

\paragraph{Diachronic corpora and temporal datasets.}
The earliest large-scale temporal corpora such as Google Books \citep{michel2011google} and Early English Books Online (EEBO) \citep{eebo, eebo_tcp} laid out the groundwork for analyzing quantitative phenomena at the interface of social sciences and humanities. However, the lack of sentence context and semantic annotations at an n-gram level preclude sentence-level inference and direct use in timely LLM fine-tuning or evaluation. Other diachronic corpora such as COHA \citep{davies2010coha}, COCA \citep{davies2008coca}, and CCOHA \citep{alatrash2020ccoha}, are valuable for capturing American English variation, but remain too small in scale to train, so as to meaningfully probe modern-scale models. Newer resources like TiC-LM \citep{li2025tic} and TemporalWiki \citep{jang2022temporalwiki} emphasize contemporary factual content rather than gradual connotative and affective evolution that characterises long-term semantic change. \CHRONOBERG complements these prior efforts by providing a literary corpus containing full-length texts with temporal metadata, facilitating both semantic and affective analysis at yearly granularity in the context of modern-day LLMs. 

\paragraph{Distributional semantics and VAD lexicons.}
The distributional hypothesis \citep{bullinaria2007semantic} forms the basis of all modern approaches to semantic change, where meaning is approximated by context and shifts in context distribution signals shifts in meaning. Count based methods using SVD \citep{levy2015svd} were superseded by predict-based embeddings in Word2vec \citep{mikolov2013word2vec}, whose temporal extension through alignment techniques like Compass-aligned distributional embeddings (CADE) \citep{bianchi2020cade} enables temporal comparisons without the instability of naive concatenation. Validation studies confirm that these methods recover documented changes, through cosine displacement \citep{hamilton2016diachronic}, change-point detection \citep{kulkarni2015statistically}, community annotation in SemEval-2020 \citep{schlechtweg2020semeval} as well as typological approaches \citep{blank1997, kay2009historicaloed, cassotti2024using}. Critically, however, these benchmarks operate on small curated word sets, which makes them unsuitable for evaluating downstream behaviour at scale. \\
The affective dimension adds a further layer of complexity. Rooted in influential factor analysis \citep{osgood1957,russell2003core}, the VAD framework decomposes word meaning into three dimensions: \textit{valence} (the positive/negative evaluative quality of a word), \textit{arousal} (the active/passive or exciting/calm tone) and \textit{dominance} (the dominant/submissive or degree of control or power the word conveys). These dimensions have been human-annotated for 45,000 contemporary English words \citep{mohammad2018vad, mohammad2025nrcvadlexiconv2}, providing fine-grained affective grounding. Diachronic embedding studies further demonstrate that such affective associations shift systematically over decades \citep{garg2018wordembeddings}, reflecting broader cultural and social change. 
Recent works incorporate affective dimensions directly within semantic change frameworks, through multidimensional evaluation schemes that connect lexical change to social science constructs such as sentiment and intensity \citep{baes2024multidimensional} and sense-level benchmarks for denotational and connotational meaning relations \citep{cassotti2026senserel}. Surveys examine how change types such as pejoration and amelioration map onto connotational dimensions \citep{desa2024survey}, while other works link semantic change to interpretable affective features \citep{oda2025improving}. Similarly, computational analysis of dehumanizing language connects affective lexical representations to social evaluation \citep{dehumanization}. As with semantic change benchmarks above, these approaches operate on curated word sets rather than corpus-scale lexicons.\\
Yet existing VAD lexicons are synchronic in nature; the scores reflect only contemporary linguistic understanding and do not account for historical semantic evolution. Consequently, the resource cannot track the changing connotations of words such as \textit{broadcast} (originally denoting the physical scattering of seeds before acquiring its modern media sense), or \textit{febrile} (which once carried strong connotations of danger and mortality before becoming a relatively neutral clinical descriptor), as modern ratings fail to capture their historical usage \citep{perc2012evolution}. \CHRONOBERG addresses this gap by contributing computationally constructed, temporally aligned VAD lexicons spanning 335,804 words across five 50-year epochs between the years 1750-2000, which are then used for sentence-level VAD annotations across the full corpus, enabling the study of affective change and temporal robustness at the scale required for LLM safety evaluation.

\paragraph{Temporal robustness of language models.}
The temporal degradation of LLMs is well-documented at the task level, with works showing that perplexity rises monotonically with temporal distance from the training window \citep{lazaridou2021mindgap} and also how this generalization gap persists across several NLP tasks despite domain-adaptive pretraining \citep{luu2022time}. A key insight from both the studies is that the gap is not merely a knowledge problem, where models simply lack recent facts, but rather a distributional one. Surface statistics, syntax and word usage all shift in ways that trained models cannot anticipate. Works on temporal adaptation \citep{rottger2021temporaladaptation} show that continual pretraining on target-period text recovers performance within that period but does not improve cross-period generalization, suggesting that temporal robustness requires evaluation resources that span long horizons rather than adjacent time windows. This has also been framed as a training pipeline problem \citep{dhingra2022time}, which argues for time-stamped corpora such as \CHRONOBERG as a prerequisite for temporally aware models.

\paragraph{Hate-speech detection and temporal bias.} 
The surface-token problem in hate-speech detection, where classifiers are triggered by the presence of a slur regardless of syntactic role or pragmatic context, was identified in \citep{davidson2017hate}, where the lexical classifier struggled to classify Tweets and sentences without explicit hate keywords. This conflation of token identity with intent is precisely the failure \CHRONOBERG has been designed to quantify. The temporal dimension compounds the problem, with works showing that classifiers trained on social-media text from one period generalize poorly even to the following year \citep{rottger2021temporaladaptation}, while others demonstrate that incorporating temporal lexical features can yield measurable improvements \citep{mcgillivray2022timedependent}. A more comprehensive study \citep{dibonaventura2025hatevolution} evaluated several language models and found that static test sets underestimate the effect of temporality; models that appear robust on fixed benchmarks fail when data drawn from a shifted distribution was used. Broader surveys on fairness and bias in hate-speech detection \citep{blodgett2020survey, rottger2021hatecheck} further establish this finding. \CHRONOBERG contributes computationally constructed temporally aligned VAD lexicons, which are then used for sentence-level VAD annotations across the dataset, making the study of the interaction between connotative drift and classifier sensitivity empirically tractable. 

\subsection{Data Source}
As outlined in our dataset construction pipeline, we now present a detailed explanation of the rationale behind our choice of data corpus. The initial step in constructing a text dataset involves identifying an appropriate data source. For temporal datasets, two primary criteria are essential: first, the availability of timestamps indicating when the data was created; second, a substantial volume of content to ensure comprehensive coverage across different time periods. While large datasets are generally beneficial for language modelling since their performance improves with the amount of training data, temporal datasets require extensive data to capture variations over time. 

In order to meet these requirements, we seek extensive text collections produced over the past centuries, easily accessible for research purposes, and accompanied by metadata detailing date of creation. Books emerged as a natural solution since they offer coherent and curated content, especially when compared to shorter form  content like news or social media posts. 

However, using books for large-scale, temporally annotated datasets presents several practical challenges: 
\begin{enumerate}
    \item Copyright Restrictions: Many books are under copyright restrictions, limiting free access to their full texts. 
    \item Digitization Requirements: To be usable, books must be available in digital formats. 
    \item Metadata Availability: Metadata, such as accurate author names and publication years, is crucial, since manually annotating hundreds of thousands of books without this information is unfeasible. 
    \item Programmatic Access: Efficient data collection necessitates programmatic interaction with the data source, such as through APIs, to download and filter relevant books in bulk. 
\end{enumerate}

We examined various online book databases, including Google Books \citep{michel2011google}, the Internet Archive, and Project Gutenberg \citep{pgh}, to assess their suitability for large-scale historical text collection. Google Books offers an extensive online library, with full-text search across a vast collection of books and metadata such as publication dates. However, the API imposes significant constraints, limiting queries to a maximum of 40 results per request. Pagination requires numerous inefficient calls, and large-scale automated retrieval is hindered by protective measures such as CAPTCHAs. Moreover, the API requires a keyword-based search and does not permit queries based solely on publication year, making systematic dataset construction difficult. 

The Internet Archive provides a large repository of digitized texts, often with richer metadata than Google Books. However, the quality and completeness of metadata  varies considerably across the entries, and publication year information is frequently missing or unreliable. Additionally, bulk access is limited by rate restrictions and heterogeneity of formats, which further complicates large-scale preprocessing.

Project Gutenberg is a widely used digital library of public domain literary works, providing unrestricted access to full-length texts. Each entry is accompanied by metadata covering attributes such as \textit{title, authors, subjects} and \textit{issue date}, as shown in Table \ref{tab:metadata_gutenberg}, together with other metadata fields available. We therefore select this as the primary resource for curating our \CHRONOBERG dataset. However, key metadata, most notably publication dates, are frequently absent or inconsistent, which poses challenges for constructing a coherent, temporally stratified dataset. The methodological details for our curation process to address this are discussed in the following section. 

\renewcommand\theadalign{bc}

\begin{table*}[h!]
\noindent
    \centering
    \begin{tabular}{>{\raggedright}p{3cm}p{7.5cm} >{\centering\arraybackslash}p{3cm}}
    \toprule
        \textbf{Attribute} & \textbf{Explanation} & \textbf{In Catalogue}  \\
    \midrule
        \textbf{ID} & A real number assigned by Project Gutenberg  to uniquely identify the eBook  & \checkmark \\ 
        \textbf{Type} & Text (\textgreater98 \%), dataset, sound, image [...] & \checkmark  \\
        \textbf{Issued} & Release date of the book & \checkmark \\
        \textbf{Title} &  The title of the book & \checkmark \\
        \textbf{Language} & The language in which the book is available & \checkmark \\
        \textbf{Authors} & All authors of the eBook & \checkmark \\
        \textbf{Subjects} & Library of Congress subject headings & \checkmark \\
        \textbf{LoCC} & Library of Congress entries & \checkmark \\
        \textbf{Bookshelves} & Hand-curated eBook collections supplemented by 64  ``Browsing'' categories which were automatically  assigned to mimic browsing in a bookstore  & \checkmark \\
        \textbf{Publisher} & The publisher of the book  & \ding{55} \\
        \textbf{License \& Rights} & Specifies the book's copyright status (e.g public domain in the USA) and, when applicable, the specific license governing its use. & \ding{55} \\
        \textbf{Downloads} & How often the book has been downloaded & \ding{55} \\
        \textbf{Birth/death dates} &  Birth and death rates of all authors and translators if available & \ding{55} \\
        \textbf{Description \& MARC520} & Description and Summary of the eBook & \ding{55} \\
        \textbf{Translators} & All translators of the eBook (if any) & \ding{55} \\
         \textbf{Datatypes} & E.g. text, HTML, ePub, PDFs, [...] & \ding{55} \\
    \bottomrule
    \end{tabular}
    \caption{Project Gutenberg metadata fields with descriptions and catalogue availability. While many bibliographic attributes are present (\checkmark), crucial information such as original publication dates is missing (\ding{55}), necessitating external inference for the construction of \CHRONOBERG. }
    \label{tab:metadata_gutenberg}
\end{table*}

\subsection{Dataset Collection and Curation}
\label{sec:datacuration}
Project Gutenberg contains some metadata inaccuracies, most notably the original year of publication often reflects the date of digitization rather than the actual release year. Accurate publication dates are critical for curating \CHRONOBERG, as our aim is to order books chronologically for further study. \\

\textbf{Metadata and External Sources.} To address this limitation, we leverage available metadata attributes such as title, author, and the author’s birth and death dates. We also queried multiple external sources such as Google Books \citep{googlebooks}, Google Search, the Library of Congress API \citep{loc}, Open Library \citep{openlibrary}, Wikipedia \citep{wikipedia}, using book titles and author lifespans to determine the correct publication year. While dates explicitly mentioned in book titles can provide additional cues when available, this method is applicable only to a limited subset of works. To assess the reliability of these tools, we manually annotated 100 books from Project Gutenberg, evenly spread between 1611 and 1912. Due to access limitations, Google Books and Google Search were excluded, and the Library of Congress API was discarded because of poor performance. 

Among the two remaining methods, shown in Table \ref{tab:pub_year_predictors}, Open Library was the most reliable, correctly estimating publication years for 49 out of 100 tested books, with a Mean Average Error (MAE) of 3.05 years and an accuracy (acc@5, i.e.~within 5 years margin) of 76\%, which falls within the acceptable tolerance range for our analysis. The Wikipedia API retrieved years for 25 books with an accuracy of 42.0\%. However, due to the limited sample sizes for Wikipedia, its reliability remains uncertain. The Open Library API provided sufficiently accurate data to support the construction of a chronological text corpus, particularly because we constrained errors within the author’s lifespan. \\

\begin{table}[t!]
\centering
\small
\resizebox{\columnwidth}{!}{%
\begin{tabular}{l c c c}
\toprule
\textbf{Scores}            & \textbf{Open Library} & \textbf{Wikipedia} & \textbf{Majority Vote} \\
\midrule
Values                     & 92    & 47    & 96 \\
Correct                    & 49    & 25    & 41 \\
Accuracy @0 (\%)           & 53.3 & 53.2 & 42.7 \\
Accuracy @5 (\%)           & 76.0 & 39.0 & 74.0 \\
Accuracy @7 (\%)           & 79.0 & 42.0 & 80.0 \\
MAE (years)                & 3.05  & 3.36  & 4.05 \\
Standard Deviation (years) & 5.20  & 7.24  & 6.56 \\
\bottomrule
\end{tabular}
}
\caption{Comparison of publication year inference methods against 100 manually annotated ground truth samples. Accuracy is reported at different tolerances, with acc@0 representing the exact year, acc@5 within 5 years and acc@7 within 7 years. Open Library emerges as the most reliable predictor with broad coverage and low error (MAE = 3.05 years), making it the most reliable stand-alone source. Wikipedia achieves limited recall and lower accuracy. Majority voting across predictors offers marginal recall gains but does not scale well, reinforcing our choice of Open Library as the default predictor.}
\label{tab:pub_year_predictors}
\end{table}

\textbf{Filtering.} The publication dates were constrained to fall within the author's year of birth and death as an additional consistency check. In case of missing information, a default lifespan of 100 years was assumed. When API methods returned multiple year estimates, the most frequently occurring year (i.e., the modal year) was selected; in the case of a tie, the first appearing year was chosen. Further filtration criteria were applied to refine \CHRONOBERG: 
 
\begin{enumerate}[label=(\roman*)]
    \item \textit{Publication Year:} Books lacking any inferrable publication year (40.7\%) were excluded. We restricted the dataset to books published between 1750 and 2000. This range balances token availability per decade and ensures linguistic consistency, as early modern English had largely been superseded by late modern English by 1750. Additionally, this timeframe allows for the analysis of historical shifts in public perception and hate-speech evolution by contrasting older texts with those from the modern era.
    \item \textit{Author Metadata:} Only books with a known author and recorded birth year were retained, leading to the exclusion of 22.9\% of volumes. To ensure plausibility, we retained only works published within the author's lifespan; for authors with unknown death years, this corresponds to a default cut-off of 100 years after birth. This step removes posthumous editions and maintains consistency.
    \item \textit{Language:} The dataset was limited to English-language texts (80.3\%)  of the total books to reduce cross-linguistic variation and maintain consistency in temporal language analysis.
    \item \textit{Translations:} Translated works (8.4\%) were removed, since their publication dates often deviate substantially from the original text, potentially distorting historical trends.
    \item \textit{Content Type:} Non-textual materials such as images and audio files (1.7\%) were excluded to preserve a purely text-based corpus.
    \item \textit{Copyright and Availability:} Only books explicitly marked as public domain in the U.S. (98.8\%) were included. Of these, 75 books lacked downloadable plaintext files in Project Gutenberg.
\end{enumerate}

With these filtering steps, we successfully annotated 25,061 books for chronological sorting out of the 73,500 books available in Project Gutenberg. \\

\begin{figure}
    \centering
    \includegraphics[width=\linewidth]{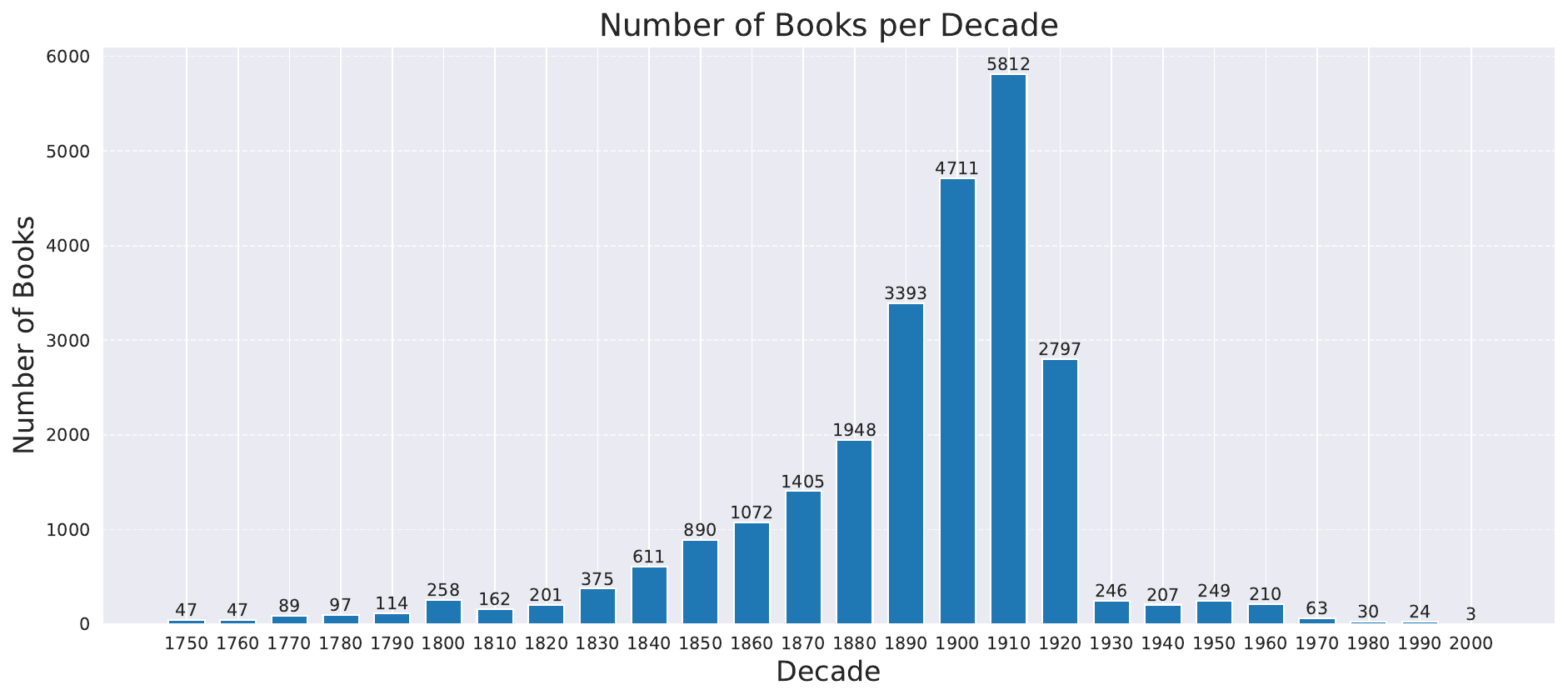}
    \caption{Temporal distribution of books in \CHRONOBERG across decades (1750 - 2000). The number of available texts grows steadily until the 1920s, likely reflecting improved literacy and publication, but drops sharply thereafter due to copyright restrictions on works published post 1920. Consequently, early 20th-century works dominate the distribution, while post-1920s coverage remains sparse.}
    \label{fig:cb_stats}
\end{figure}

\textbf{Topic Distribution.}
The distribution of books and tokens in \CHRONOBERG across decades is uneven, as illustrated in Figure \ref{fig:cb_stats}, reflecting the availability of texts in Project Gutenberg. The number of books increases steadily up to the 1920s, a trend likely fueled by population growth, educational expansion, and economic development. A sharp decline follows, primarily due to copyright restrictions, with most texts published post-1929 remaining under copyright, and consequently being unavailable in Project Gutenberg. As a result, early 20th-century works are well represented, while post-1920 works are underrepresented. This is especially evident for the 2000s, which contain only the year 2000, producing a notably low count of books and tokens for that decade. 

\begin{table}[t!]
\small
\centering
\begin{tabular}{l c c}
    \toprule
    \multicolumn{3}{c}{\emph{\textbf{Subjects}}} \\
    Subject & Books & \% \\
    \midrule
    Fiction & 5998 & 23.9 \\
    History & 3012 & 12.0 \\
    Juvenile fiction & 2368 & 9.4 \\
    Social Life and Customs & 1450 & 5.8 \\
    19th century & 1218 & 4.9 \\
    England & 1154 & 4.6 \\
    United States & 1052 & 4.2 \\
    Great Britain & 872 & 3.5 \\
    Description and Travel & 830 & 3.3 \\
    Conduct of Life & 695 & 2.8 \\
    \bottomrule
    \\[-0.5em]
    \toprule
    \multicolumn{3}{c}{\emph{\textbf{Bookshelves}}} \\
    Bookshelf & Books & \% \\
    \midrule
    American Bestsellers 1895--1923 & 308 & 7.2 \\
    Science Fiction & 285 & 6.7 \\
    Children's Fiction 1895--1923 & 282 & 6.6 \\
    Children's Series 1895--1923 & 207 & 4.9 \\
    Children's Literature & 203 & 4.8 \\
    World War I & 196 & 4.6 \\
    US Civil War & 192 & 4.5 \\
    Historical Fiction & 186 & 4.4 \\
    Humor & 100 & 2.4 \\
    Native America & 98 & 2.3 \\
    \bottomrule
\end{tabular}
\caption{Top 10 most frequent subjects and bookshelf topics in \CHRONOBERG. Proportion of books assigned to each category are shown in \% relative to the total number of books with at least one entry in the same category. ``Subjects'' are derived from Library of Congress headings, while ``Bookshelves'' are mostly from automatically assigned ``browsing'' categories. Note that multiple subject headings and bookshelves may be assigned to a single book.}
\label{tab:subjects}
\end{table}

\begin{table*}[t]
\centering
\setlength{\tabcolsep}{6pt}
\renewcommand{\arraystretch}{1.2}

\begin{tabular}{cc}
\begin{minipage}{0.49\linewidth}
\centering
\textbf{1750--1799}\\[3pt]
\begin{tabular}{l r}
\toprule
Author & Books \\
\midrule
Gibbon, Edward & 14 \\
Schiller, Friedrich & 7 \\
Paine, Thomas & 7 \\
Wollstonecraft, Mary & 7 \\
Stanhope, Philip & 7 \\
\bottomrule
\end{tabular}
\end{minipage}
&
\begin{minipage}{0.49\linewidth}
\centering
\textbf{1800--1849}\\[3pt]
\begin{tabular}{l r}
\toprule
Author & Books \\
\midrule
Bulwer-Lytton, Edward & 88 \\
Scott, Walter & 44 \\
Marryat, Frederick & 40 \\
Dickens, Charles & 39 \\
Cooper, James Fenimore & 33 \\
\bottomrule
\end{tabular}
\end{minipage}
\\ \\

\begin{minipage}{0.45\linewidth}
\centering
\textbf{1850--1899}\\[3pt]
\begin{tabular}{l r}
\toprule
Author & Books \\
\midrule
Twain, Mark & 104 \\
Ballantyne, Robert Michael & 88 \\
Henty, George Alfred & 74 \\
Alger, Horatio Jr. & 72 \\
Fenn, George Manville & 66 \\
\bottomrule
\end{tabular}
\end{minipage}
&
\begin{minipage}{0.45\linewidth}
\centering
\textbf{1900--1949}\\[3pt]
\begin{tabular}{l r}
\toprule
Author & Books \\
\midrule
Baum, Lyman Frank & 73 \\
Stratemeyer, Edward & 59 \\
Barbour, Ralph Henry & 57 \\
Oppenheim, Edward Phillips & 55 \\
Wells, Carolyn & 54 \\
\bottomrule
\end{tabular}
\end{minipage}
\\ \\

\begin{minipage}{0.45\linewidth}
\centering
\textbf{1950--2000}\\[3pt]
\begin{tabular}{l r}
\toprule
Author & Books \\
\midrule
Leinster, Murray & 17 \\
Duellman, William Edward & 11 \\
Dick, Philip K. & 10 \\
Kjelgaard, Jim & 10 \\
Norton, Andre & 10 \\
\bottomrule
\end{tabular}
\end{minipage}
&
\begin{minipage}{0.45\linewidth}
\centering
\textbf{All time}\\[3pt]
\begin{tabular}{l r}
\toprule
Author & Books \\
\midrule
Twain, Mark & 118 \\
Bulwer-Lytton, Edward & 104 \\
Ballantyne, Robert Michael & 88 \\
Fenn, George Manville & 87 \\
Henty, George Alfred & 86 \\
\bottomrule
\end{tabular}
\end{minipage}
\end{tabular}

\caption{\CHRONOBERG contains more than 20,000 authors, out of which the most represented ones are shown and grouped by historical period (1750–2000). Book counts indicate the number of works attributed to each author within the dataset and illustrate shifts in author prominence over time.}
\label{tab:most_represented_authors}
\end{table*}

Beyond temporal coverage, Project Gutenberg metadata also provides insights into \mbox{\CHRONOBERG's} thematic composition. Table~\ref{tab:subjects} presents the ten most common subjects and bookshelf topics, reflecting the historical context in which the books were written. Fiction, in its various forms, is the most prominent genre in \CHRONOBERG. Many works also reflect social and historical contexts followed by works addressing social issues and major historical events such as World War I and U.S. Civil War. Historical works covering earlier periods are also well-represented. In contrast, children’s literature constitutes a small portion of the dataset: the three bookshelf categories related to children's literature account for only 692 books, forming a relatively niche subject.

The subjects used to categorize the books are drawn from the Library of Congress Subject Headings, whereas the categories for bookshelves are derived from a mixture of hand-curated eBook collections and automatically assigned ``browsing'' categories. Multiple subject headings or bookshelves can be assigned to a single book. The completeness of the metadata varies, where only 22 books in \CHRONOBERG lack subject headings, but around 83\% not having any bookshelf assigned. These were excluded from the counts shown in Table \ref{tab:subjects}.

We also examine the authors represented in \CHRONOBERG over time. Out of the 20k authors in \CHRONOBERG, the most represented authors are shown in Table \ref{tab:most_represented_authors}, illustrating shifts in author prominence across 250 years. While several prolific writers (e.g., Dickens, Schiller, Twain) appear prominently, the corpus is largely shaped by authors who were popular in their time, but are rarely read today. This distribution reflects typical biases of large-scale digitized historical corpora, capturing the broader landscape of historical print culture, including many commercially popular authors whose work may better represent everyday linguistic usage of their period.

\subsection{Investigating the effect of the number of neighbours on VAD measures}
\label{app:section_neighbors}

\begin{figure*}[t]
    \centering
    \includegraphics[width=0.32\textwidth]{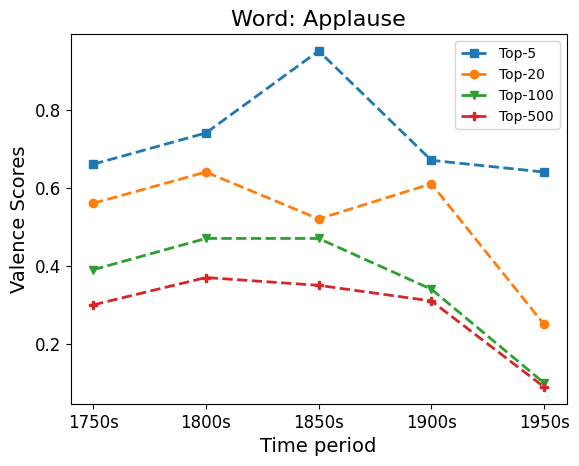}
    \includegraphics[width=0.32\textwidth]{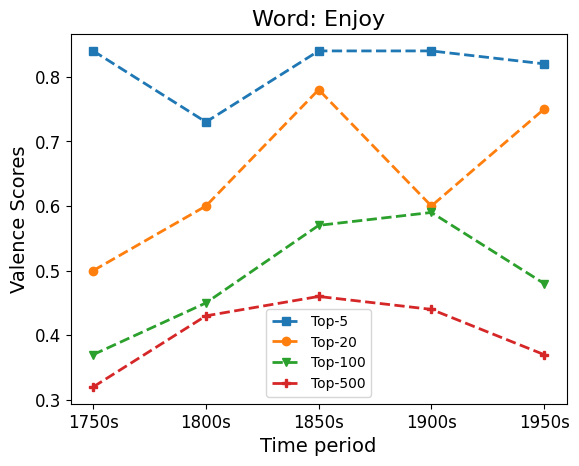}
    \includegraphics[width=0.32\textwidth]{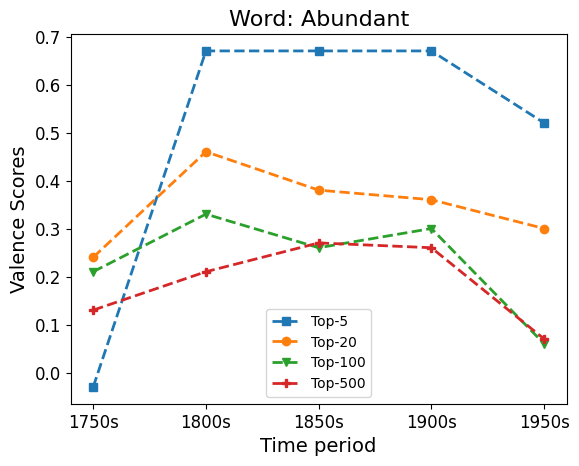}
    \caption{Visualizing the variance in the valence scores across the different time intervals. We vary the number of the Top-K neighbours to compute the affective valence scores for each word. Top-5 neighbours lead to strong bias, whereas top-500 neighbours lead to neutrality.}
    \label{fig:neighbor}
\end{figure*}

As discussed in Section 2.3 of the main text, the number of neighbours used to compute lexical scores is a crucial hyperparameter when constructing temporally aligned VAD lexicons. This choice directly affects the resulting VAD scores, where too few neighbours can introduce strong biases, while too many can produce overly neutral scores. Figure \ref{fig:neighbor} illustrates this effect across the different eras for a select set of examples.

\begin{figure*}[t]
    \centering
    \includegraphics[width=0.49\linewidth] {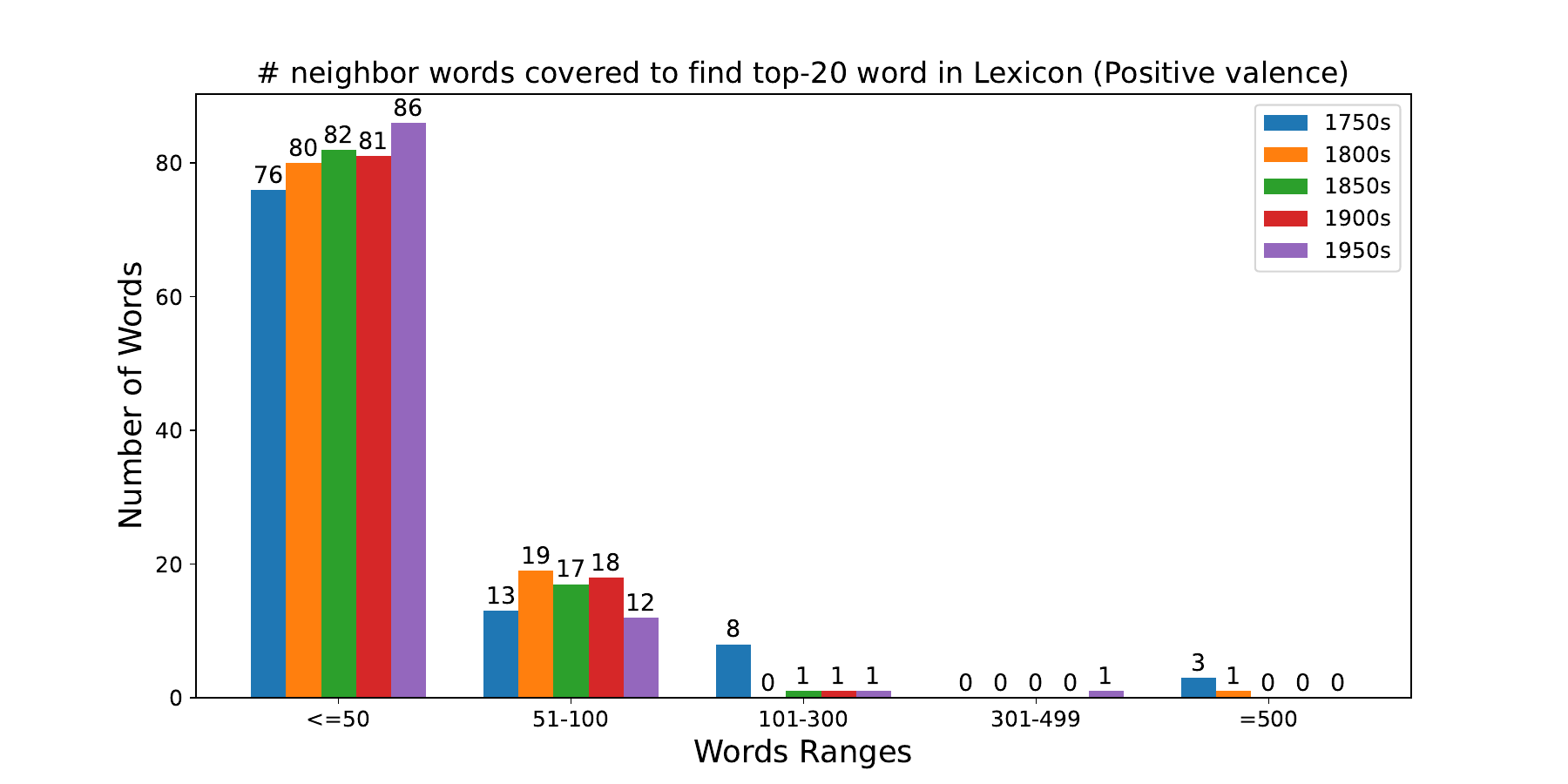}
\includegraphics[width=0.49\linewidth] {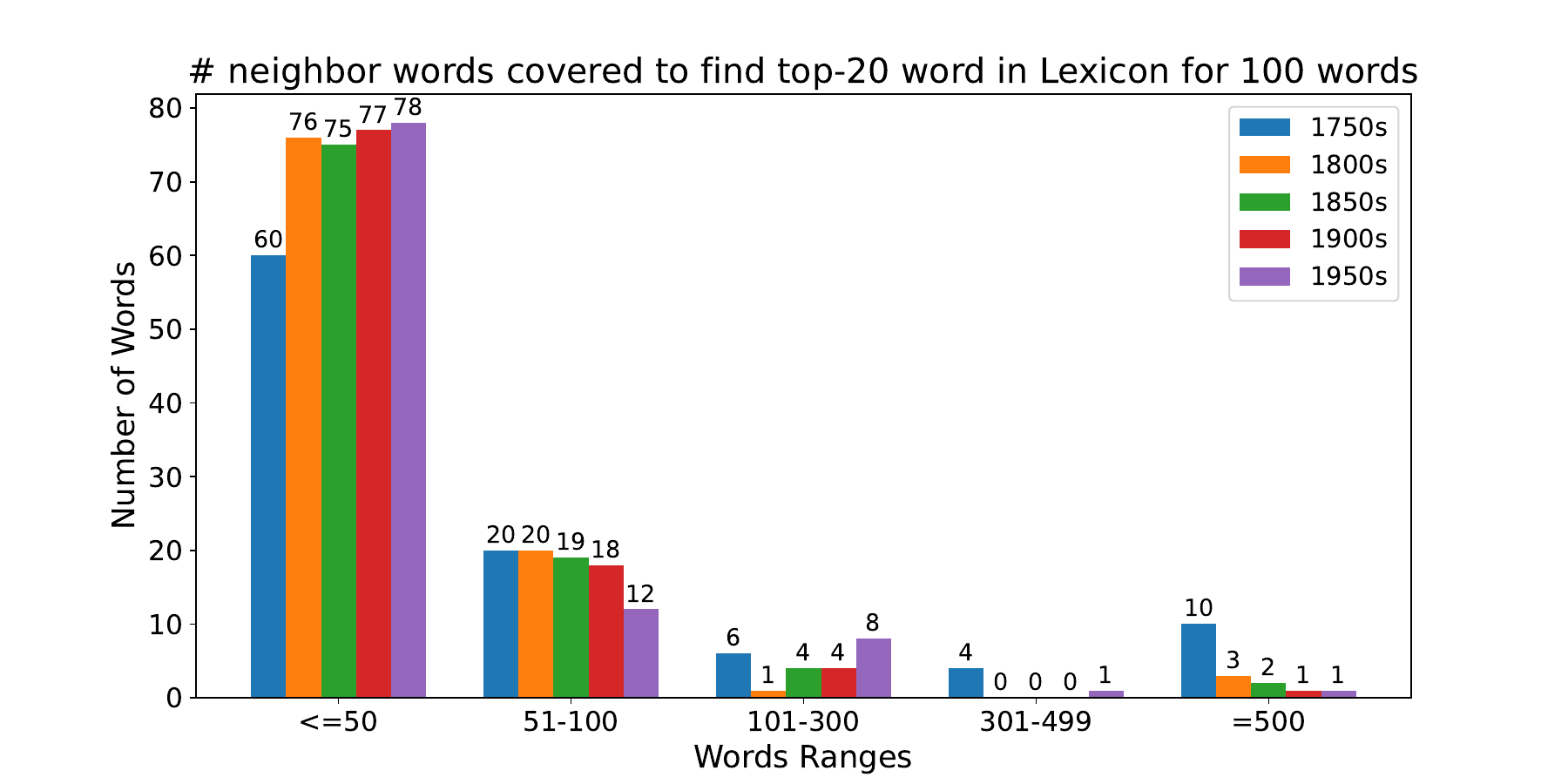}
        \caption{Visualizing the number of neighbours needs to be traversed to determine a VAD score by identifying the top-20 words in NRC VAD Lexicons. For most words, top-50 neighbours are sufficient while top-100 nearest neighbours provide a reliable upper bound covering all words. }
    \label{fig:sample_top_100}
\end{figure*}

From Figure \ref{fig:neighbor} it is observable that using the top-20 neighbours is typically adequate. However, because \CHRONOBERG's lexicons are based on the contemporary NRC VAD lexicon, not all top-20 neighbours in historical sentences may have corresponding NRC VAD scores. To address this, we conducted a small experiment, where we randomly sampled 100 words and measured how many nearest neighbours must be traversed to identify the 20 words present in the NRC VAD lexicon. The results in Figure \ref{fig:sample_top_100} show that for most words, considering the top 50 neighbours is sufficient to obtain 20 valid scores, and the top 100 neighbours provide a reliable upper bound covering all words. Based on this analysis, we adopt the top-100 neighbours as the standard for computing VAD scores in \CHRONOBERG, ensuring robust and consistent affective annotations. 
\subsection{Further examples for semantic shifts of words in \CHRONOBERG}
\label{app:dia_examples}
\begin{figure}[t!]
    \centering
    \includegraphics[width=\columnwidth]{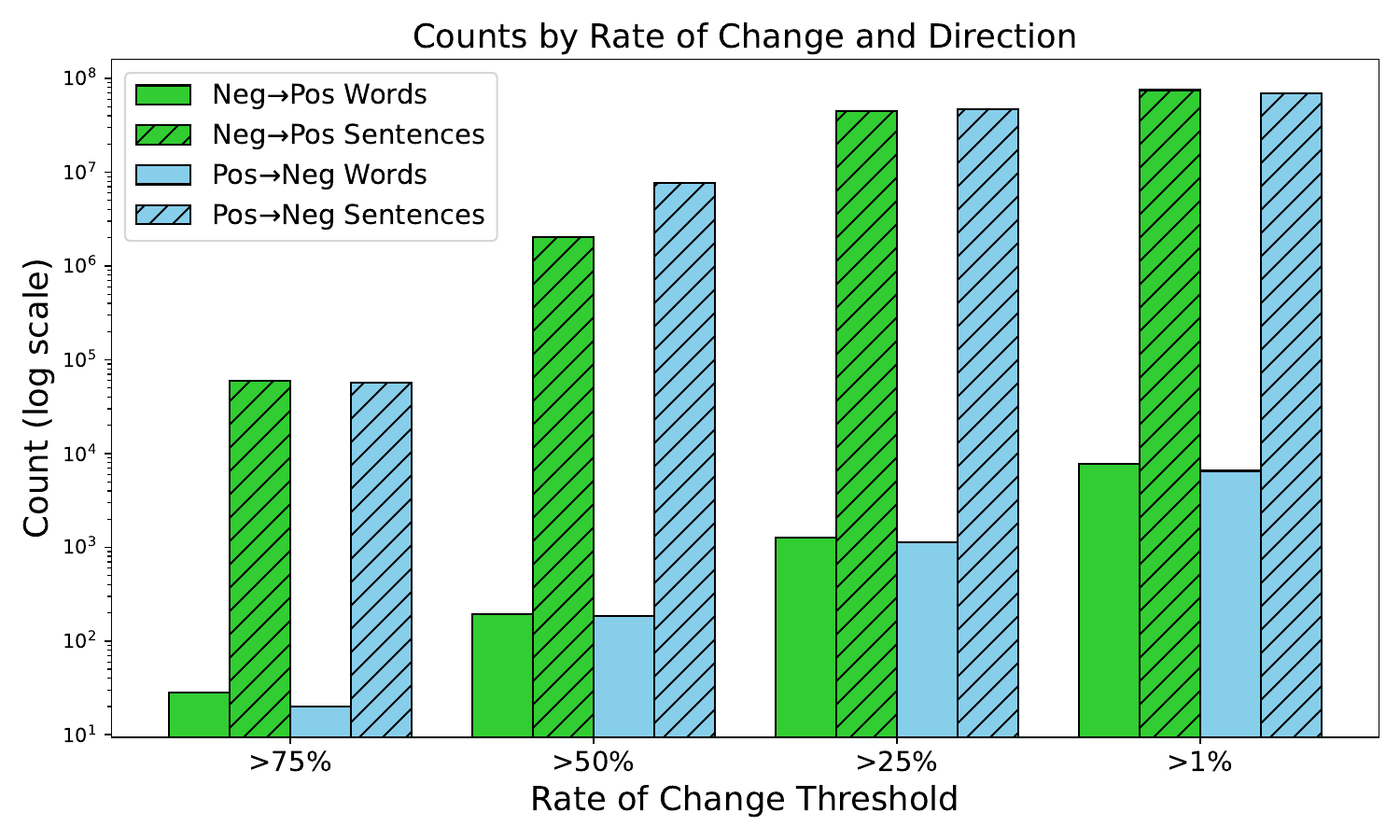}
    \caption{Distribution of valence shifts between consecutive 50-year intervals in \CHRONOBERG. Bars show the count of words and sentences changing from positive to negative and vice versa, across different thresholds of rate of change. While most words remain effectively stable, thousands of samples exhibit substantial shifts in both directions.}
    \label{fig:shifts_valence}
\end{figure}

Figure~\ref{fig:shifts_valence} reports the distribution of valence shifts between the selected epochs. We present more qualitative examples that illustrate the transition of affective connotations, both from positive to negative and vice versa in \Cref{tab:dia_shifts}. For instance, “sanctimonious”, which once conveyed genuine holiness, has shifted to denote a hypocritical display of moral superiority. Likewise, weird, originally associated with the supernatural and unearthly, now predominantly means “odd,” “strange,” or “bizarre.” Another compelling case is “depressive”, which evolved from a neutral or even positive association of “pressing down” to a word carrying strongly negative emotional connotations. 
 
\begin{table*}[h!]
\centering
\small 
\setlength{\tabcolsep}{3.5pt} 

\begin{minipage}{0.48\linewidth}
\centering
\begin{tabular}{l c c c c c}
    \toprule
    & \multicolumn{5}{c}{\emph{Positive $\rightarrow$ Negative}} \\
    \bfseries Words & \bfseries 1750s & \bfseries 1800s & \bfseries 1850s & \bfseries 1900s & \bfseries 1950s \\
    \midrule
    asylum & 0.27 & -0.24 & -0.54 & -0.52 & -0.65 \\ 
    coronary & 0.17 & -0.13 & -0.22 & -0.15 & -0.68 \\ 
    depressive & 0.3 & -0.96 & -0.56 & -0.65 & -0.54 \\ 
    germs & 0.15 & 0.26 & -0.14 & -0.55 & -0.61 \\ 
    heartbreak & 0.18 & -0.6 & -0.69 & -0.74 & -0.78 \\ 
    \bottomrule
\end{tabular}
\end{minipage}
\hfill 
\begin{minipage}{0.48\linewidth}
\centering
\begin{tabular}{l c c c c c}
    \toprule
    & \multicolumn{5}{c}{\emph{Negative $\rightarrow$ Positive}} \\
    \bfseries Words & \bfseries 1750s & \bfseries 1800s & \bfseries 1850s & \bfseries 1900s & \bfseries 1950s \\
    \midrule
    bloomers & 0.01 & -0.05 & 0.27 & 0.18 & 0.38 \\ 
    destiny & -0.54 & 0.06 & 0.32 & 0.11 & 0.55 \\ 
    dunk & 0.42 & 0.15 & None & -0.18 & 0.19 \\ 
    febrile & -0.58 & -0.53 & -0.66 & -0.54 & 0.33 \\ 
    infatuation & -0.66 & -0.63 & -0.52 & -0.35 & 0.40 \\ 
    \bottomrule
\end{tabular}
\end{minipage}
\caption{We present additional examples illustrating how the valence scores of words have shifted across centuries. The left column shows words that carried positive connotations historically but have since acquired negative ones, while the right column shows the inverse: words that were once negatively connoted but have taken on positive connotations over time. }
\label{tab:dia_shifts}
\end{table*}

\begin{table*}[h!]
\centering
\small
\setlength{\tabcolsep}{3.5pt} 
\begin{minipage}{0.48\linewidth}
\centering
\begin{tabular}{l c c c c c}
    \toprule
    & \multicolumn{5}{c} {\emph{Predominantly Negative}}\\
    \bfseries Words & \bfseries 1750s & \bfseries 1800s & \bfseries 1850s & \bfseries 1900s & \bfseries 1950s \\
    \midrule
    afraid & -0.35 & -0.34 & -0.09 & -0.27 & -0.39\\
    angered & -0.20 & -0.58 & -0.66 & -0.66 & -0.69\\
    annihilation & -0.49 & -0.68 & -0.62 & -0.73 & -0.61\\
    bankruptcy & -0.33 & -0.37 & -0.46 & -0.55 & -0.46\\
    betray & -0.52 & -0.44 & -0.51 & -0.49 & -0.66\\
    chaos & -0.27 & -0.33 & -0.23 & -0.46 & -0.57\\
    stabbed & -0.48 & -0.68 & -0.74 & -0.60 & -0.56\\
    strangulation & -0.60 & -0.47 & -0.67 & -0.69 & -0.26\\
    suicidal & -0.27 & -0.75 & -0.70 & -0.71 & -0.59\\
    \bottomrule
\end{tabular}
\end{minipage}
\hfill
\begin{minipage}{0.48\linewidth}
\centering
\begin{tabular}{l c c c c c}
    \toprule
    & \multicolumn{5}{c} {\emph{Predominantly Positive}} \\
    \bfseries Words & \bfseries 1750s & \bfseries 1800s & \bfseries 1850s & \bfseries 1900s & \bfseries 1950s \\
    \midrule
    abundant & 0.28 & 0.37 & 0.40 & 0.40 & 0.41\\
    enjoy & 0.52 & 0.60 & 0.62 & 0.63 & 0.86\\
    hugs & -0.28 & 0.34 & 0.56 & 0.56 & 0.40\\
    laughter & 0.08 & 0.33 & 0.49 & 0.49 & 0.75\\
    liking & 0.35 & 0.07 & 0.12 & 0.28 & 0.21\\
    lucky & -0.14 & 0.05 & 0.33 & 0.30 & 0.29\\
    marvel & 0.03 & 0.56 & 0.63 & 0.77 & 0.84\\
    merry & 0.56 & 0.71 & 0.76 & 0.79 & 0.54\\
    respectful & 0.45 & 0.42 & 0.41 & 0.40 & 0.62\\
    \bottomrule
\end{tabular}
\end{minipage}
\caption{Our VAD lexicons also capture words whose connotations have remained stable, largely positive or negative, over the 250 years of literary texts in \CHRONOBERG, without undergoing any dramatic semantic shift. }
\label{tab:dia_shifts_2}
\end{table*}

\begin{table*}[t]
\centering
\small
\setlength{\tabcolsep}{2.8pt}
\begin{tabular}{l *{13}{c}}
\toprule
& \multicolumn{13}{c}{\emph{Positive $\rightarrow$ Negative}} \\
\cmidrule(l){2-14}
\textbf{Word} & \textbf{1750s} & \textbf{1770s} & \textbf{1790s} & \textbf{1810s} & \textbf{1830s} & \textbf{1850s} & \textbf{1870s} & \textbf{1890s} & \textbf{1910s} & \textbf{1930s} & \textbf{1950s} & \textbf{1970s} & \textbf{1990s} \\
\midrule
asylum   & $0.16$ & $0.06$  & $0.16$  & $-0.01$ & $-0.20$ & $-0.50$ & $-0.49$ & $-0.54$ & $-0.52$ & $-0.54$ & $-0.46$ & $-0.10$ & $-0.10$ \\
germs    & $0.30$ & $0.28$  & $0.21$  & $0.28$  & $0.31$  & $0.17$  & $0.15$  & $-0.54$ & $-0.52$ & $-0.68$ & $-0.53$ & $-0.53$ & $-0.06$ \\
homeless & $0.20$ & $-0.23$ & $-0.44$ & $-0.53$ & $-0.66$ & $-0.65$ & $-0.64$ & $-0.66$ & $-0.63$ & $-0.49$ & $-0.52$ & $-0.52$ & $-0.20$ \\
punk     & $0.17$ & $-0.31$ & $-0.16$ & $-0.16$ & $-0.16$ & $-0.21$ & $-0.21$ & $-0.16$ & $-0.12$ & $-0.35$ & $-0.14$ & $-0.11$ & $-0.11$ \\
weird    & $0.20$ & $0.20$  & $0.05$  & $-0.11$ & $0.00$  & $-0.32$ & $-0.29$ & $-0.26$ & $-0.36$ & $-0.47$ & $-0.44$ & $-0.17$ & $-0.17$ \\
\bottomrule
\end{tabular}
\caption{Valence trajectories for words undergoing a positive-to-negative shift, derived from lexicons built on Word2Vec models trained on 20-year temporal slices. Words that shift in the 50-year models show comparable transitions here, but the 20-year splits reveal a finer-grained progression of the valence scores than the coarser counterpart.}
\label{tab:shifts_20yr}
\end{table*}
Conversely, we also observe cases of semantic shift from negative to positive. Words such as infatuation, destiny, tweak, and repertoire exemplify this trend. “Infatuation” has moved from its earlier sense of “making foolish” to its modern meaning of intense admiration. Similarly, “tweak”, once meaning “to pluck or pinch,” has broadened to signify the act of making small adjustments. We have also reported several instances of words that were predominantly positive or negative across the time interval of 250 years, as shown in \Cref{tab:dia_shifts_2}. 

\subsection{External Validation of Temporal VAD lexicons}
\label{app:validation_app}

\paragraph{Method.}
To assess whether \CHRONOBERG's temporal VAD lexicons capture documented semantic
shifts, we apply a two-threshold confidence heuristic to each word's valence
trajectory across five 50-year epochs. Words with fewer than three observed values
are excluded. For each remaining word, we compute consecutive per-epoch deltas
between adjacent observed values, skipping missing epochs without interpolation.
A step is \textit{significant} if $|\Delta_i| > 0.1$. We also compute the overall
net shift $\Delta_\text{net} = v_\text{last} - v_\text{first}$ independently.
Confidence tiers are assigned as follows: \colorbox{highgreen}{\textbf{H}}igh requires at least two significant steps \textit{and} $|\Delta_\text{net}| > 0.2$: the trajectory shows a clear, multi-step directional shift. \colorbox{medyellow}{\textbf{M}}edium is assigned when local movement is present but the net shift is weak ($|\Delta_\text{net}| \leq 0.2$), or only a single significant step is observed: signal is detectable but not unambiguous. Finally, \colorbox{lowred}{\textbf{L}}ow is assigned when the shift is invisible to the
VAD lexicons, either because no step exceeds the noise floor or because only sub-threshold drift accumulates. \textit{Stable} is assigned when no meaningful signal is present on either measure. Direction is assigned following this procedure:  if
$|\Delta_\text{net}| > 0.2$, direction follows the sign of the net shift directly; if a linear regression over the observed epoch values yields $R^2 \geq 0.3$ and $|\Delta_\text{net}| > 0.1$, direction follows the sign of the regression slope; otherwise direction is determined by the dominant sign among significant consecutive steps, with net shift sign as tiebreaker. Table~\ref{tab:vad_validation} in the main paper reports aggregated results; full per-word trajectories appear in Tables~\ref{tab:app_full_a}--\ref{tab:app_full_b3} below.

\paragraph{Evaluative Changes.}
Of the 40 in-scope evaluative words, 23 are correctly detected by the thresholded VAD lexicons. 8 \colorbox{highgreen}{\textbf{H}}-tier pejorations agree with the expected direction, confirming that clear, dominant evaluative shifts produce reliable VAD signals. Representative examples include \textit{awful} ($-0.20{\to}-0.66$), \textit{Bastard} ($+0.11{\to}-0.48$), and \textit{propaganda} ($-0.06{\to}-0.28$). All 4 \colorbox{highgreen}{\textbf{H}}-tier amelioration words are also correctly detected (\textit{awesome}: $-0.78{\to}-0.12$; \textit{nice}: $+0.18{\to}+0.47$; \textit{sophisticated}: $+0.03{\to}+0.27$; \textit{Witch}: $-0.30{\to}-0.03$). Medium-tier detections capture a further
11 words where the VAD signal is present but weaker, including the ameliorations \textit{Minx} ($-0.31{\to}-0.12$), \textit{terrific} ($-0.61{\to}-0.29$), \textit{Vixen} ($-0.20{\to}-0.06$), and \textit{maverick} ($-0.12{\to}+0.02$). \\

The 16 misrepresentations fall into two interpretable categories:
(a) \textit{Direction mismatch ($\ddagger$):} Eleven words show corpus-level movement in the wrong direction. Words such as \textit{bully} ($-0.54{\to}-0.18$) and \textit{egregious} ($-0.64{\to}-0.36$) appear to ameliorate in book text because their dominant literary senses pull aggregate neighbourhood valence upward, masking the pejorative shift that occurred primarily in spoken and informal registers. Similarly, \textit{nervous} ($-0.52{\to}-0.35$), \textit{demagogue} ($-0.52{\to}-0.40$), and \textit{Pervert} ($-0.44{\to}-0.34$) show small net positive shifts driven by oscillation around their already-negative baseline. \textit{Batty} ($-0.27{\to}-0.01$) rises toward neutral in the literary corpus as the standard British English sense dominates, while the slur sense remained confined to informal registers. For Amelioration, \textit{guy} ($+0.17{\to}-0.09$) is the sole miss: despite its modern colloquial positivity, the word carries slightly negative literary connotations across the full 250-year window, reflecting its historical association with grotesquely dressed effigies. Notably, \textit{gay/gays} ($+0.12{\to}+0.52$) shows a strong positive trajectory (consistent with reclamation rather than pejoration). \\
(b) \textit{Register gap ($\star$):} Five words are assigned \colorbox{lowred}{\textbf{L}}ow because their pejorated senses were too infrequent in edited book text to produce measurable valence movement. \textit{Fatal} ($-0.62{\to}-0.67$) was already strongly negative before 1750, leaving no
trajectory to detect. \textit{Weasel} ($-0.05{\to}+0.00$) and \textit{Pig} ($+0.02{\to}+0.05$) remain near-neutral because the animal senses dominate published text. \textit{Queen} ($+0.42{\to}+0.35$) and \textit{Fairy} ($+0.39{\to}+0.46$) stay consistently positive throughout, as the regal and mythological senses respectively outweigh the pejorated senses in \CHRONOBERG. \textit{Sanction} ($+0.18{\to}+0.19$) is N/A: its two opposing meanings coexist and cancel, making VAD direction assessment inapplicable.

\begin{table*}[t!]
\centering
\footnotesize
\setlength{\tabcolsep}{3pt}
\renewcommand{\arraystretch}{1}
\begin{tabular}{l l r r r r r l l c}
\toprule
\textbf{Word} & \textbf{Type}
& $v_{1750}$ & $v_{1800}$ & $v_{1850}$ & $v_{1900}$ & $v_{1950}$
& \textbf{VAD trend} & \textbf{Tier} & \textbf{Agr.} \\
\midrule
\multicolumn{10}{l}{\textit{Evaluative --- Pejoration}} \\
\midrule
awful         & Pejoration & $-0.20$ & $-0.48$ & $-0.55$ & $-0.68$ & $-0.66$ & Pejoration   & \colorbox{highgreen}{\textbf{H}} & \checkmark \\
bastard       & Pejoration & $+0.11$ & $+0.25$ & $-0.05$ & $-0.35$ & $-0.48$ & Pejoration   & \colorbox{highgreen}{\textbf{H}} & \checkmark \\
fag           & Pejoration & $-0.15$ & $+0.09$ & $-0.12$ & $-0.16$ & $-0.51$ & Pejoration   & \colorbox{highgreen}{\textbf{H}} & \checkmark \\
hermaphrodite & Pejoration & $+0.15$ & $-0.06$ & $+0.06$ & $+0.00$ & $-0.44$ & Pejoration   & \colorbox{highgreen}{\textbf{H}} & \checkmark \\
moron         & Pejoration & $+0.06$ & ---     & $+0.22$ & $-0.59$ & $-0.26$ & Pejoration   & \colorbox{highgreen}{\textbf{H}} & \checkmark \\
propaganda    & Pejoration & $-0.06$ & $+0.00$ & $+0.14$ & $-0.24$ & $-0.28$ & Pejoration   & \colorbox{highgreen}{\textbf{H}} & \checkmark \\
slut          & Pejoration & $-0.08$ & $+0.00$ & $-0.33$ & $-0.56$ & $-0.50$ & Pejoration   & \colorbox{highgreen}{\textbf{H}} & \checkmark \\
twit          & Pejoration & $+0.35$ & $-0.46$ & $-0.51$ & $-0.55$ & $-0.25$ & Pejoration   & \colorbox{highgreen}{\textbf{H}} & \checkmark \\
batty         & Pejoration & $-0.27$ & $-0.10$ & $+0.04$ & $+0.14$ & $-0.01$ & Amelioration & \colorbox{highgreen}{\textbf{H}} & \texttimes \\
\midrule
bitch         & Pejoration & $+0.11$ & $+0.11$ & $+0.16$ & $+0.19$ & $-0.21$ & Pejoration   & \colorbox{medyellow}{\textbf{M}} & \checkmark \\
dyke          & Pejoration & $-0.09$ & $-0.14$ & $-0.14$ & $-0.28$ & $-0.33$ & Pejoration   & \colorbox{medyellow}{\textbf{M}} & \checkmark \\
prick         & Pejoration & $-0.31$ & $-0.18$ & $-0.30$ & $-0.45$ & $-0.41$ & Pejoration   & \colorbox{medyellow}{\textbf{M}} & \checkmark \\
queer         & Pejoration & $-0.24$ & $+0.00$ & $+0.16$ & $-0.14$ & $-0.27$ & Pejoration   & \colorbox{medyellow}{\textbf{M}} & \checkmark \\
bender        & Pejoration & $+0.05$ & $+0.06$ & $+0.00$ & $-0.20$ & $-0.01$ & Pejoration   & \colorbox{medyellow}{\textbf{M}} & \checkmark \\
faggot        & Pejoration & $-0.06$ & $-0.08$ & $+0.00$ & $-0.02$ & $-0.15$ & Pejoration   & \colorbox{medyellow}{\textbf{M}} & \checkmark \\
nigger        & Pejoration & $+0.08$ & $-0.05$ & $-0.32$ & $-0.36$ & $+0.03$ & Pejoration   & \colorbox{medyellow}{\textbf{M}} & \checkmark \\
\midrule
bully         & Pejoration & $-0.54$ & $-0.62$ & $-0.55$ & $-0.51$ & $-0.18$ & Amelioration & \colorbox{medyellow}{\textbf{M}} & \texttimes \\
demagogue     & Pejoration & $-0.52$ & $-0.11$ & $-0.44$ & $-0.16$ & $-0.40$ & Amelioration & \colorbox{medyellow}{\textbf{M}} & \texttimes \\
dick          & Pejoration & $-0.10$ & $+0.05$ & $+0.16$ & $+0.10$ & $-0.01$ & Amelioration & \colorbox{medyellow}{\textbf{M}} & \texttimes \\
egregious     & Pejoration & $-0.64$ & $-0.63$ & $-0.67$ & $-0.62$ & $-0.36$ & Amelioration & \colorbox{medyellow}{\textbf{M}} & \texttimes \\
gay/gays      & Pejoration & $+0.12$ & $+0.07$ & $+0.10$ & $+0.52$ & ---     & Amelioration & \colorbox{medyellow}{\textbf{M}} & \texttimes \\
homo          & Pejoration & $+0.00$ & ---     & $-0.14$ & $-0.01$ & $+0.04$ & Amelioration & \colorbox{medyellow}{\textbf{M}} & \texttimes \\
nervous       & Pejoration & $-0.52$ & $-0.34$ & $-0.45$ & $-0.41$ & $-0.35$ & Amelioration & \colorbox{medyellow}{\textbf{M}} & \texttimes \\
pervert       & Pejoration & $-0.44$ & $-0.62$ & $-0.79$ & $-0.79$ & $-0.34$ & Amelioration & \colorbox{medyellow}{\textbf{M}} & \texttimes \\
pig           & Pejoration & $+0.02$ & $-0.13$ & $-0.13$ & $-0.09$ & $+0.05$ & Amelioration & \colorbox{medyellow}{\textbf{M}} & \texttimes \\
\midrule
fatal         & Pejoration & $-0.62$ & $-0.72$ & $-0.66$ & $-0.65$ & $-0.67$ & Stable       & \colorbox{lowred}{\textbf{L}} & \texttimes \\
weasel        & Pejoration & $-0.05$ & $-0.06$ & $-0.08$ & $-0.08$ & $+0.00$ & Stable       & \colorbox{lowred}{\textbf{L}} & \texttimes \\
queen         & Pejoration & $+0.42$ & $+0.45$ & $+0.42$ & $+0.41$ & $+0.35$ & Stable       & \colorbox{lowred}{\textbf{L}} & \texttimes \\
bent          & Pejoration & $-0.06$ & $-0.01$ & $-0.07$ & $-0.12$ & $-0.04$ & Stable       & \colorbox{lowred}{\textbf{L}} & \texttimes \\
fairy         & Pejoration & $+0.39$ & $+0.46$ & $+0.45$ & $+0.47$ & $+0.46$ & Stable       & \colorbox{lowred}{\textbf{L}} & \texttimes \\
\bottomrule
\end{tabular}
\caption{Full validation results for in-scope evaluative words (Part~1a):
Pejoration. Valence scores are per-epoch estimates from \CHRONOBERG's temporal VAD lexicons. Confidence tiers: \colorbox{highgreen}{\textbf{H}}igh = strong multi-step signal ($\geq$2 steps, $|\Delta_\text{net}|>0.2$); \colorbox{medyellow}{\textbf{M}}edium = signal present but weak or single-step; \colorbox{lowred}{\textbf{L}}ow = shift invisible to VAD lexicons (register gap or sub-threshold drift). \checkmark~= VAD direction agrees with expected annotated change; \texttimes~= does not agree. `---'~= no data for that epoch.}
\label{tab:app_full_a}
\end{table*}
 
\begin{table*}[t!]
\centering
\footnotesize
\setlength{\tabcolsep}{3pt}
\renewcommand{\arraystretch}{1}
\begin{tabular}{l l r r r r r l l c}
\toprule
\textbf{Word} & \textbf{Type}
& $v_{1750}$ & $v_{1800}$ & $v_{1850}$ & $v_{1900}$ & $v_{1950}$
& \textbf{VAD trend} & \textbf{Tier} & \textbf{Agr.} \\
\midrule
\multicolumn{10}{l}{\textit{Evaluative --- Amelioration}} \\
\midrule
awesome       & Amelioration & ---     & $-0.78$ & $-0.40$ & $-0.31$ & $-0.12$ & Amelioration & \colorbox{highgreen}{\textbf{H}} & \checkmark \\
nice          & Amelioration & $+0.18$ & $+0.54$ & $+0.68$ & $+0.57$ & $+0.47$ & Amelioration & \colorbox{highgreen}{\textbf{H}} & \checkmark \\
sophisticated & Amelioration & $+0.03$ & $-0.29$ & $-0.34$ & $+0.23$ & $+0.27$ & Amelioration & \colorbox{highgreen}{\textbf{H}} & \checkmark \\
witch         & Amelioration & $-0.30$ & $-0.15$ & $-0.30$ & $-0.16$ & $-0.03$ & Amelioration & \colorbox{highgreen}{\textbf{H}} & \checkmark \\
\midrule
minx          & Amelioration & $-0.31$ & $-0.25$ & $-0.18$ & $-0.28$ & $-0.12$ & Amelioration & \colorbox{medyellow}{\textbf{M}} & \checkmark \\
terrific      & Amelioration & $-0.61$ & $-0.55$ & $-0.51$ & $-0.49$ & $-0.29$ & Amelioration & \colorbox{medyellow}{\textbf{M}} & \checkmark \\
vixen         & Amelioration & $-0.20$ & $+0.16$ & $-0.18$ & $-0.06$ & $-0.06$ & Amelioration & \colorbox{medyellow}{\textbf{M}} & \checkmark \\
maverick      & Amelioration & $-0.12$ & ---     & $+0.67$ & $+0.07$ & $+0.02$ & Amelioration & \colorbox{medyellow}{\textbf{M}} & \checkmark \\
\midrule
guy           & Amelioration & $+0.17$ & $+0.04$ & $+0.02$ & $+0.15$ & $-0.09$ & Pejoration   & \colorbox{highgreen}{\textbf{H}} & \texttimes \\
\midrule
\multicolumn{10}{l}{\textit{Evaluative --- Contronym}} \\
\midrule
sanction      & Contronym    & $+0.18$ & $+0.22$ & $+0.23$ & $+0.30$ & $+0.19$ & Stable       & \colorbox{medyellow}{\textbf{M}} & \texttimes \\
\bottomrule
\end{tabular}
\caption{Full validation results for evaluative words (Part~1b):
Amelioration and Contronym. See Table~\ref{tab:app_full_a} for column definitions.}
\label{tab:app_full_a2}
\end{table*}

\paragraph{Non-evaluative Changes.}
Out of the 43 non-evaluative words, 8 correctly show flat valence, validating that the VAD lexicons do not generate spurious evaluative signals for non-evaluative changes. The words showing valence movement offer analytically interesting insights into how Non-evaluative change produces connotative drift.

\textit{Metaphor:} Words such as \textit{film} ($-0.20{\to}+0.14$) and \textit{network} ($+0.00{\to}+0.17$) drift positive as their technological senses become dominant in later epochs. \textit{Record} ($+0.30{\to}+0.11$), \textit{stream} ($+0.21{\to}+0.09$), and \textit{prop} ($+0.19{\to}-0.09$) decline in positivity as their associations with recordings, digital media, and theatrical props displace earlier neutral or positive uses. \textit{Virus} ($-0.34{\to}-0.57$) shows a consistent negative drift as the pathological sense consolidates across the corpus window.

\textit{Narrowing:} \textit{Skyline} ($+0.63{\to}+0.02$) shows the sharpest single-epoch drop ($-0.59$) as its meaning narrowed from a general horizon term to an urban architectural concept with more neutral connotations.

\textit{Subjectification:} \textit{Actually} ($-0.01{\to}+0.27$) and \textit{starting} ($-0.13{\to}+0.16$) drift positive as they grammaticalise into affiliative discourse markers. \textit{Hopefully} ($+0.02{\to}-0.14$) drifts negative as its subjective ``I hope'' sense became associated with uncertainty and tentativeness.

\textit{Multiple coexisting senses:} \textit{Check} ($-0.35{\to}+0.21$) shows a dramatic positive shift as the confirmatory sense grew relative to the restraint/verification sense. \textit{Stroke} ($-0.40{\to}-0.05$) ameliorates
as positive senses (caress, brushstroke, stroke of luck) are represented alongside the medical sense.

\textit{SemEval-2020 Stable control words \citep{schlechtweg2020semeval}:} Of the 24 control words, 9 are correctly classified as Stable (\textit{bag}, \textit{donkey}, \textit{fiction}, \textit{plane}, \textit{player}, \textit{risk}, \textit{tip}, \textit{tree}, \textit{word}). The remaining 15 show apparent movement, which is consistent with \CHRONOBERG's 250-year span: SemEval-2020 defines stability over a short two-timepoint window, so genuine long-horizon drift may account for some cases. The two strongest false positives are \textit{quilt} ($-0.15{\to}+0.34$) and \textit{multitude} ($-0.25{\to}+0.04$), both of which show upward drift likely driven by changes in genre distribution across epochs rather than genuine semantic change.

\begin{table*}[t!]
\centering
\footnotesize
\setlength{\tabcolsep}{3pt}
\renewcommand{\arraystretch}{1}
\begin{tabular}{l l r r r r r l}
\toprule
\textbf{Word} & \textbf{Type}
& $v_{1750}$ & $v_{1800}$ & $v_{1850}$ & $v_{1900}$ & $v_{1950}$
& \textbf{VAD trend} \\
\midrule
\multicolumn{8}{l}{\textit{Non-evaluative --- Metaphor}} \\
\midrule
attack    & Metaphor & $-0.29$ & $-0.24$ & $-0.33$ & $-0.29$ & $-0.33$ & Stable   \\
bug       & Metaphor & $-0.31$ & $-0.29$ & $-0.35$ & $-0.33$ & $-0.24$ & Stable   \\
edge      & Metaphor & $-0.08$ & $-0.04$ & $-0.05$ & $-0.07$ & $-0.13$ & Stable   \\
land      & Metaphor & $+0.18$ & $+0.13$ & $+0.18$ & $+0.16$ & $+0.14$ & Stable   \\
platform  & Metaphor & $+0.03$ & $+0.06$ & $+0.12$ & $+0.10$ & $+0.08$ & Stable   \\
\midrule
bit       & Metaphor & $+0.14$ & $-0.13$ & $-0.15$ & $-0.13$ & $-0.19$ & Shifting \\
broadcast & Metaphor & $+0.09$ & $+0.13$ & $+0.06$ & $-0.05$ & $+0.07$ & Shifting \\
cell      & Metaphor & $-0.06$ & $+0.01$ & $-0.01$ & $-0.16$ & $-0.04$ & Shifting \\
cloud     & Metaphor & $+0.04$ & $-0.08$ & $-0.05$ & $-0.14$ & $+0.06$ & Shifting \\
film      & Metaphor & $-0.20$ & $-0.04$ & $+0.06$ & $+0.14$ & $+0.14$ & Shifting \\
mouse     & Metaphor & $-0.17$ & $-0.07$ & $-0.05$ & $-0.03$ & $+0.02$ & Shifting \\
network   & Metaphor & $+0.00$ & $-0.03$ & $+0.08$ & $+0.01$ & $+0.17$ & Shifting \\
prop      & Metaphor & $+0.19$ & $+0.18$ & $+0.14$ & $-0.01$ & $-0.09$ & Shifting \\
rag       & Metaphor & $+0.06$ & $-0.07$ & $-0.05$ & $-0.09$ & $-0.12$ & Shifting \\
record    & Metaphor & $+0.30$ & $+0.32$ & $+0.20$ & $+0.15$ & $+0.11$ & Shifting \\
stab      & Metaphor & $-0.48$ & $-0.49$ & $-0.50$ & $-0.39$ & $-0.49$ & Shifting \\
stream    & Metaphor & $+0.21$ & $+0.08$ & $+0.13$ & $+0.14$ & $+0.09$ & Shifting \\
tablet    & Metaphor & $+0.12$ & $+0.05$ & $-0.12$ & $-0.08$ & $+0.01$ & Shifting \\
thump     & Metaphor & $-0.19$ & $-0.38$ & $-0.44$ & $-0.42$ & $-0.28$ & Shifting \\
virus     & Metaphor & $-0.34$ & $-0.52$ & $-0.44$ & $-0.38$ & $-0.57$ & Shifting \\
web       & Metaphor & $-0.01$ & $-0.01$ & $+0.03$ & $+0.02$ & $-0.10$ & Shifting \\
\midrule
\multicolumn{8}{l}{\textit{Non-evaluative --- Narrowing}} \\
\midrule
car       & Narrowing & $+0.21$ & $+0.16$ & $+0.16$ & $+0.13$ & $+0.10$ & Shifting \\
diet      & Narrowing & $+0.13$ & $+0.00$ & $+0.19$ & $+0.22$ & $+0.13$ & Shifting \\
estate    & Narrowing & $+0.34$ & $+0.35$ & $+0.33$ & $+0.19$ & $+0.31$ & Shifting \\
skyline   & Narrowing & ---     & $+0.63$ & $+0.04$ & $+0.05$ & $+0.02$ & Shifting \\
\bottomrule
\end{tabular}
\caption{Full validation results for non-evaluative and control words (Part 2a). The ``Type'' column indicates whether the change in the meaning of the words falls into one of these categories: ``Metaphor'', ``Narrowing'', ``Generalization'', ``Subjectification'', or ``Multiple coexisting senses'' (``Polysemy'' or ``Domain shift''). ``VAD Trend'' being ``Stable'' indicates valence correctly stays flat; ``Shifting'' indicates an unexpected directional shift. Within each subcategory, stable words precede shifting words. \emph{(Continued in Parts 2b--2c.)}}
\label{tab:app_full_b1}
\end{table*}

\begin{table*}[t!]
\centering
\footnotesize
\setlength{\tabcolsep}{3pt}
\renewcommand{\arraystretch}{1}
\begin{tabular}{l l r r r r r l}
\toprule
\textbf{Word} & \textbf{Type}
& $v_{1750}$ & $v_{1800}$ & $v_{1850}$ & $v_{1900}$ & $v_{1950}$
& \textbf{VAD trend} \\
\midrule
\multicolumn{8}{l}{\textit{Non-evaluative --- Generalization}} \\
\midrule
head      & Generalization & $-0.02$ & $-0.04$ & $+0.02$ & $-0.02$ & $-0.05$ & Stable   \\
\midrule
circle    & Generalization & $+0.01$ & $+0.13$ & $+0.08$ & $+0.07$ & $+0.08$ & Shifting \\
lass      & Generalization & $+0.19$ & $+0.33$ & $+0.22$ & $+0.11$ & $-0.10$ & Shifting \\
\midrule
\multicolumn{8}{l}{\textit{Non-evaluative --- Subjectification}} \\
\midrule
major     & Subjectification & $+0.09$ & $+0.10$ & $+0.09$ & $+0.10$ & $+0.06$ & Stable   \\
\midrule
actually  & Subjectification & $-0.01$ & $+0.09$ & $+0.21$ & $+0.10$ & $+0.27$ & Shifting \\
headed    & Subjectification & $-0.15$ & $+0.03$ & $-0.04$ & $+0.04$ & $-0.03$ & Shifting \\
hopefully & Subjectification & $+0.02$ & $+0.09$ & $+0.25$ & $-0.01$ & $-0.14$ & Shifting \\
promise   & Subjectification & $+0.25$ & $+0.34$ & $+0.27$ & $+0.26$ & $+0.37$ & Shifting \\
started   & Subjectification & $-0.19$ & $+0.01$ & $+0.02$ & $-0.04$ & $+0.01$ & Shifting \\
starting  & Subjectification & $-0.13$ & $-0.04$ & $-0.05$ & $+0.09$ & $+0.16$ & Shifting \\
\midrule
\multicolumn{8}{l}{\textit{Non-evaluative --- Multiple coexisting senses}} \\
\midrule
computer  & Domain/Technology & ---     & ---     & $+0.12$ & $+0.16$ & $+0.19$ & Stable   \\
\midrule
call      & Polysemy          & $-0.01$ & $+0.12$ & $+0.22$ & $+0.21$ & $+0.19$ & Shifting \\
calls     & Polysemy          & $+0.04$ & $-0.03$ & $-0.03$ & $-0.05$ & $+0.12$ & Shifting \\
check     & Polysemy          & $-0.35$ & $-0.31$ & $-0.25$ & $+0.21$ & $+0.21$ & Shifting \\
democrat  & Domain/Technology & $+0.02$ & $-0.08$ & $+0.03$ & $+0.04$ & $+0.07$ & Shifting \\
monitor   & Polysemy          & $+0.02$ & $+0.22$ & $-0.05$ & $-0.11$ & $+0.13$ & Shifting \\
privacy   & Domain/Technology & $+0.21$ & $+0.09$ & $+0.15$ & $+0.14$ & $+0.25$ & Shifting \\
stroke    & Polysemy          & $-0.40$ & $-0.34$ & $-0.24$ & $-0.11$ & $-0.05$ & Shifting \\
\bottomrule
\end{tabular}
\caption{Full validation results for non-evaluative words, continued (Part 2b). See Table~\ref{tab:app_full_b1} for column definitions. \emph{(Continued in Part 2c.)}}
\label{tab:app_full_b2}
\end{table*}

\begin{table*}[t!]
\centering
\footnotesize
\setlength{\tabcolsep}{3pt}
\renewcommand{\arraystretch}{1}
\begin{tabular}{l l r r r r r l}
\toprule
\textbf{Word} & \textbf{Type}
& $v_{1750}$ & $v_{1800}$ & $v_{1850}$ & $v_{1900}$ & $v_{1950}$
& \textbf{VAD trend} \\
\midrule
\multicolumn{8}{l}{\textit{SemEval-2020 Stable}} \\
\midrule
bag           & Stable            & $+0.01$ & $+0.04$ & $+0.04$ & $+0.05$ & $+0.05$ & Stable   \\
donkey        & Stable            & $+0.01$ & $-0.01$ & $+0.08$ & $+0.03$ & $+0.07$ & Stable   \\
fiction       & Stable            & $+0.36$ & $+0.37$ & $+0.33$ & $+0.30$ & $+0.40$ & Stable   \\
plane         & Stable            & $-0.02$ & $-0.02$ & $+0.03$ & $+0.05$ & $-0.01$ & Stable   \\
player        & Stable            & $+0.30$ & $+0.26$ & $+0.31$ & $+0.30$ & $+0.29$ & Stable   \\
risk          & Stable            & $-0.26$ & $-0.33$ & $-0.27$ & $-0.37$ & $-0.35$ & Stable   \\
tip           & Stable            & $-0.08$ & $-0.14$ & $-0.10$ & $-0.07$ & $-0.14$ & Stable   \\
tree          & Stable            & $+0.11$ & $+0.06$ & $+0.11$ & $+0.10$ & $+0.06$ & Stable   \\
word          & Stable            & $+0.15$ & $+0.14$ & $+0.05$ & $+0.09$ & $+0.16$ & Stable   \\
\midrule
ball          & Stable            & $-0.20$ & $+0.03$ & $+0.09$ & $+0.03$ & $+0.08$ & Shifting \\
chairman      & Stable            & $+0.11$ & $+0.24$ & $+0.20$ & $+0.21$ & $+0.20$ & Shifting \\
contemplation & Stable            & $+0.10$ & $+0.37$ & $+0.35$ & $+0.34$ & $+0.37$ & Shifting \\
face          & Stable            & $-0.14$ & $+0.00$ & $+0.05$ & $+0.00$ & $+0.00$ & Shifting \\
gas           & Stable            & $-0.12$ & $+0.01$ & $-0.03$ & $-0.06$ & $-0.17$ & Shifting \\
graft         & Stable            & $+0.10$ & $+0.15$ & $+0.12$ & $-0.64$ & $+0.15$ & Shifting \\
lane          & Stable            & $+0.06$ & $+0.17$ & $+0.20$ & $+0.20$ & $+0.11$ & Shifting \\
multitude     & Stable            & $-0.25$ & $-0.13$ & $-0.06$ & $-0.10$ & $+0.04$ & Shifting \\
ounce         & Stable            & $+0.15$ & $+0.06$ & $+0.06$ & $+0.05$ & $-0.02$ & Shifting \\
part          & Stable            & $+0.04$ & $+0.06$ & $+0.15$ & $+0.13$ & $+0.15$ & Shifting \\
pin           & Stable            & $-0.11$ & $-0.13$ & $-0.02$ & $-0.04$ & $+0.03$ & Shifting \\
quilt         & Stable            & $-0.15$ & $+0.18$ & $+0.16$ & $+0.15$ & $+0.34$ & Shifting \\
relationship  & Stable            & $+0.47$ & $+0.37$ & $+0.42$ & $+0.37$ & $+0.28$ & Shifting \\
savage        & Stable            & $-0.48$ & $-0.54$ & $-0.56$ & $-0.56$ & $-0.63$ & Shifting \\
twist         & Stable            & $-0.14$ & $-0.13$ & $-0.11$ & $-0.20$ & $-0.25$ & Shifting \\
\bottomrule
\end{tabular}
\caption{Full validation results for control words (Part 2c).
Words from SemEval-2020 \citep{schlechtweg2020semeval} annotated as ``Stable'' serve as a control group. Of the 24 words, 9 are correctly classified as Stable and 15 show apparent valence movement; some of the latter may reflect genuine long-horizon drift below SemEval's short two-timepoint detection threshold rather than classifier noise. See Table~\ref{tab:app_full_b1} for column definitions.}
\label{tab:app_full_b3}
\end{table*}

\section{Analysis of Hate Speech and Harmful Language}
\label{sec:hatespeech}

This appendix section provides further information on how suitable hate-detection tools were identified and benchmarked, how they were applied on \CHRONOBERG, and contains additional examples.

\subsection{Identifying Suitable Hate-detection Tools with HateCheck}
We provide further insights into the choices underlying the main body's hate-speech detection pipeline to contextualize harmful language in \CHRONOBERG. In total, we have considered nine different modern hate-speech detection tools: Pysentimiento toolkit \citep{pysentimiento}, Google Perspective API \citep{lees2022perspectiveapi}, Facebook RoBERTa \citep{liu2019roberta}, as well as 6 most popular Hugging Face models when filtering for the keyword ``hate''. It is worth noting that all nine tools are built upon one of the two popular transformer
architectures BERT \citep{bert} and RoBERTa \citep{liu2019roberta}. 

Consequently, we evaluated these tools on the HateCheck benchmark \citep{rottger2021hatecheck}, a suite of functional tests for hate-speech detection. As shown in Table \ref{tab:full_hatecheck_performance}, many existing hate-speech detection tools performed no better than random guessing. RoBERTa \citep{liu2019roberta} was the only tool that stood out, demonstrating consistently strong performance, achieving high recall, while the remaining tools exhibited substantial limitations in either recall, precision, or overall reliability. Notably, Perspective API \citep{lees2022perspectiveapi} achieved exceptionally high precision, making it particularly viable for curating a subset of potentially hateful sentences, despite its limited coverage.

\begin{table*}[h!]
\centering
\begin{minipage}[t]{0.48\textwidth}
\centering
\begin{tabular}{ccccc}
\toprule
Score $\in$ & Sentences & TP & FP \\
\midrule
{[0.0, 0.1)}   & 2,411,275 & - & - \\
{[0.1, 0.2)}   & 360,085   & - & - \\
{[0.2, 0.3)}   & 228,128   & - & - \\
{[0.3, 0.4)}   & 148,038   & - & - \\
{[0.4, 0.5)}   & 99,557    & - & - \\
{[0.5, 0.6)}   & 68,540    & 84 & 16 \\
{[0.6, 0.7)}   & 22,213    & 95 & 5 \\
{[0.7, 0.8)}   & 5,116     & 99 & 1 \\
{[0.8, 0.9)}   & 470       & 100 & 0 \\
{[0.9, 1.0]}   & 11        & 11 & 0 \\
\bottomrule
\end{tabular}
\captionof*{table}{(a) Distribution and precision across score intervals}
\end{minipage}
\hfill
\begin{minipage}[t]{0.48\textwidth}
\centering
\begin{tabular}{ccc}
\toprule
Score $\geq$ & Sentences & P \\
\midrule
0.0 & 3,343,433 & - \\
0.1 & 932,158   & - \\
0.2 & 572,073   & - \\
0.3 & 343,945   & - \\
0.4 & 195,907   & - \\
0.5 & 96,350  & 87.4\% \\
0.6 & 27,810    & 95.8\% \\
0.7 & 5,597     & 99.1\% \\
0.8 & 481       & 100\% \\
0.9 & 11        & 100\% \\
\bottomrule
\end{tabular}
\captionof*{table}{(b) Precision and coverage at varying thresholds}
\end{minipage}
\caption{Perspective API score analysis for sentences flagged as
hateful by RoBERTa.}
\label{tab:persp_threshold}
\end{table*}
To combine their strengths, we adopted a two-stage pipeline in the main body based on these HateCheck observations. First, we use RoBERTa \citep{liu2019roberta} to flag a broad set of potentially hateful sentences. These sentences are then filtered by the Perspective API \citep{lees2022perspectiveapi} to reduce false positives. This approach balances scalability and precision, addressing RoBERTa’s over-sensitivity and the Perspective API’s limited coverage.

\begin{table*}[t!]
\centering
\begin{tabular}{lcccc}
\toprule
\textbf{Model} & \textbf{Acc} & \textbf{F1} & \textbf{P} & \textbf{R} \\
\midrule
Perspective API \citep{lees2022perspectiveapi} & 0.578 & 0.559 & \textbf{0.993} & 0.389 \\
pysentimiento \citep{pysentimiento} & 0.521 & 0.527 & 0.820 & 0.388 \\
Facebook's RoBERTa \citep{liu2019roberta} & \textbf{0.956} & \textbf{0.968} & 0.963 & \textbf{0.973}\\
English Abusive MuRIL \citep{das2022muril} & 0.491 & 0.558 & 0.694 & 0.466 \\
BERT HateXplain \citep{mathew2021hatexplain} & 0.384 & 0.270 & 0.730 & 0.165\\
DehateBERT Mono English \citep{aluru2020deep} & 0.425 & 0.351 & 0.784 & 0.226 \\
IMSyPP Hate Speech \citep{kraljnovak2022imsypp} & 0.750 & 0.826 & 0.790 & 0.866 \\
Twitter RoBERTa Large Hate \citep{antypas2023supertweeteval} & 0.615 & 0.640 & 0.898 & 0.497\\
DistilRoBERTa Hateful Speech \citep{distilroberta2023} & 0.568 & 0.652 & 0.730 & 0.590 \\
\bottomrule
\end{tabular}
\caption{Performances of models on full HateCheck test set. The best values are indicated in bold.}
\label{tab:full_hatecheck_performance}
\end{table*}
\subsection{Sentences Containing Potential Hate in \CHRONOBERG}

We present a detailed analysis of Perspective API scores for sentences flagged as hateful by the RoBERTa model for \CHRONOBERG. In Table~\ref{tab:persp_threshold}(a), the distribution of these sentences across score intervals is shown, along with manual annotations of 100 sampled sentences per range to estimate the precision. No manual revision was conducted for scores below 0.5, as the [0.5, 0.6) range already yielded 16 false positives, indicating a substantial drop in precision. In consequence, lower thresholds seem to be impractical for reliable hate-speech filtering. 
\begin{table*}[]
\noindent
    \centering
  \resizebox{\linewidth}{!}{%
  \begin{tabular}{>{\raggedright}p{1.cm}p{10.6cm}ccccp{1cm}c}
  \toprule
  \multirow{2}{*}{\textbf{YEAR}} & \multirow{2}{*}{\textbf{Sentences}} & \multicolumn{4}{c}{\textbf{Hate-Check Models}} & \textbf{Valence} &
  \textbf{Affective} \\
  & & \textbf{RoBERTA+Persp} & \textbf{OpenAI} &\textbf{Qwen}  &\textbf{Mistral} & \textbf{Score} & \textbf{Sentiment} \\
  \midrule
  \midrule

       1800s & We has slaves too; We has n*ggers to a stand-still. & \textcolor{red}{\faFlag} & \textcolor{red}{\faFlag} &
  \textcolor{red}{\faFlag} & \textcolor{red}{\faFlag} & -0.37 & \faThumbsODown\\
       1850s & I hate women. & \textcolor{red}{\faFlag} & \textcolor{red}{\faFlag} & \textcolor{red}{\faFlag} & \textcolor{red}{\faFlag} &
  -0.40 & \faThumbsODown\\
       1900s & You never want to take a n*gger into your conferences. & \textcolor{red}{\faFlag} & \textcolor{red}{\faFlag} &
  \textcolor{red}{\faFlag} & \textcolor{red}{\faFlag} & -0.18 & \faThumbsODown\\
       1900s & Kill every black bastard befo mornin! & \textcolor{red}{\faFlag} & \textcolor{red}{\faFlag} & \textcolor{red}{\faFlag} &
  \textcolor{red}{\faFlag} & -0.36 & \faThumbsODown\\
       1900s & The Bhutanese women are the ugliest specimens of femininity I have ever seen. & \textcolor{red}{\faFlag} &
  \textcolor{red}{\faFlag} & \textcolor{red}{\faFlag} & \textcolor{red}{\faFlag} & 0.45 & \faThumbsODown\\
       1950s & I hate chicago, i hate americans! & \textcolor{red}{\faFlag} & \textcolor{red}{\faFlag} & \textcolor{red}{\faFlag} &
  \textcolor{red}{\faFlag} & -0.40 & \faThumbsODown\\
       1950s & I hate the germans! & \textcolor{red}{\faFlag} & \textcolor{red}{\faFlag} & \textcolor{red}{\faFlag} & \textcolor{red}{\faFlag}
   & -0.40 & \faThumbsODown\\
  \midrule
       1750s & A man may play with decency; but if he games, he is disgraced. & \color{green}\ding{51} & \color{green}\ding{51} &
  \color{green}\ding{51} & \color{green}\ding{51} & -0.68 & \faThumbsODown \\
       1750s & Defamation and calumny never attack, where there is no weak place; they magnify, but they do not create. &
  \color{green}\ding{51} & \color{green}\ding{51} & \color{green}\ding{51} & \color{green}\ding{51} & -0.76 & \faThumbsODown \\
       1750s & Thou traitor, hie away; By all my stars I thou enviest Tom Thumb & \color{green}\ding{51} & \textcolor{red}{\faFlag} &
  \textcolor{red}{\faFlag} & \textcolor{red}{\faFlag} & -0.69 & \faThumbsODown \\
       1800s & Who the beggar was that i killed & \color{green}\ding{51} & \color{green}\ding{51} & \textcolor{red}{\faFlag} &
  \color{green}\ding{51} & -0.60 & \faThumbsODown\\
       1800s & What hatred she distills! & \color{green}\ding{51} & \color{green}\ding{51} & \textcolor{red}{\faFlag} &
  \textcolor{red}{\faFlag} & -0.72 & \faThumbsODown \\
       1850s & The piece was stupid beyond expression & \color{green}\ding{51} & \color{green}\ding{51} & \color{green}\ding{51} &
  \color{green}\ding{51} & -0.57 & \faThumbsODown \\
       1950s & So it is a hell of women, is it? & \color{green}\ding{51} & \color{green}\ding{51} & \textcolor{red}{\faFlag} &
  \color{green}\ding{51} & -0.35 & \faThumbsODown\\
  \midrule
       1750s & The conversation at supper was very gay. & \textcolor{red}{\faFlag} & \color{green}\ding{51} & \color{green}\ding{51} &
  \color{green}\ding{51} & 0.37 & \faThumbsOUp\\
       1750s & In my way home to my tent, I saw a faggot lying in the way, & \textcolor{red}{\faFlag} & \textcolor{red}{\faFlag} &
  \textcolor{red}{\faFlag} & \textcolor{red}{\faFlag} & 0.05 & \faThumbsOUp\\
       1850s & Religion prescribes obedience. & \textcolor{red}{\faFlag} & \color{green}\ding{51} & \color{green}\ding{51} &
  \color{green}\ding{51} & 0.08 & \faThumbsOUp\\
       
       1900s & I may cut you out of my gold expedition, if you get gay. & \textcolor{red}{\faFlag} & \textcolor{red}{\faFlag} &
  \textcolor{red}{\faFlag} & \textcolor{red}{\faFlag} & 0.17 & \faThumbsOUp\\
  \bottomrule
  \end{tabular}%
  }
    \caption{Extended illustrative examples. Comparison of sentence-level classifications in \CHRONOBERG across time intervals using LLM-based hate-check tools (RoBERTa+Perspective API, OpenAI, Qwen, and Mistral) {(\textcolor{red}{\faFlag}- Hate, \color{green}{\ding{51}- Non-hate})} and valence-based scoring (\faThumbsODown- Negative, \faThumbsOUp-  Positive sentiment).
    We grouped the instances based on model agreement. The first group of rows illustrates cases where all tools collectively classify an instance as harmful. The group of rows 2-3 shows instances where they disagree, classifying them as either positive or negative. While VAD lexicons provide interpretable complementary signals, we recognize that harmful texts are inherently subjective; thus, we do not regard them as definitive solutions to LLM misclassification but as potential tools to enhance LLM performance. Note, we have partially masked offensive N-words with asterisks. }
    \label{tab:hate_speech}
\end{table*}
Table~\ref{tab:persp_threshold}(b) complements this by showing how varying the Perspective API threshold affects both the number of flagged sentences and the estimated precision. We extrapolated precision estimates from the interval-level annotations, due to the infeasibility of reviewing all 3.3 million samples. Notably, 2.4 million sentences, making up 72.1\%, that were labelled as hateful by RoBERTa scored below 0.1 in Perspective API, indicating a high false positive rate. Despite this, RoBERTa exhibited strong recall during evaluation and likely also captured a majority of hateful sentences, albeit imprecisely. Using a threshold of 0.7 with Perspective API resulted in a highly precise subset of 5,597 sentences. 

\begin{figure}[h!]
    \centering
    \includegraphics[width=\linewidth]{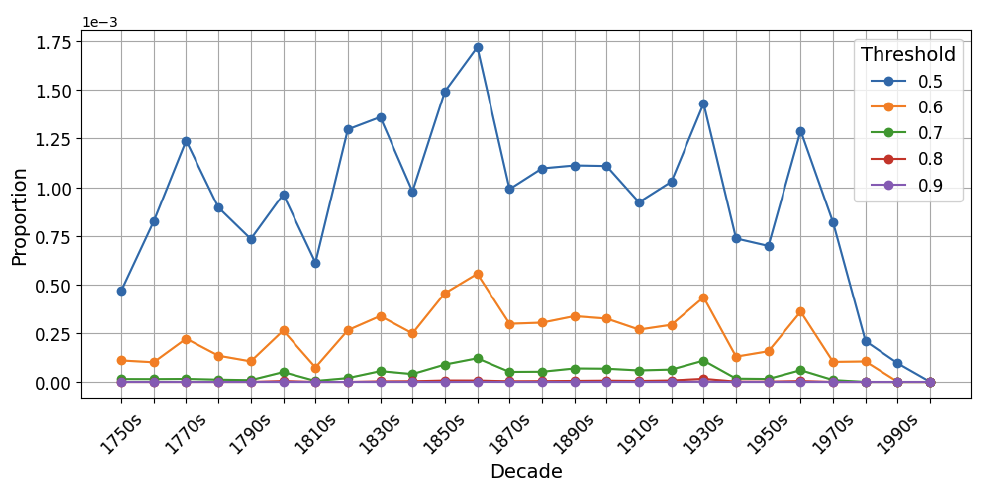}
    \caption{Proportion of hateful sentences over time for different Perspective API
thresholds}
    \label{fig:hateful_persp}
\end{figure}

We observed that while varying the Perspective API threshold influences the volume of hate detected across decades, as seen in Figure \ref{fig:hateful_persp}, the overall temporal patterns remain similar and are retained. In comparison with the distribution as flagged by RoBERTa, more substantial differences are revealed, particularly in earlier historical periods, whereas Perspective API consistently detects less hate across all thresholds. 

\subsection{Further qualitative examples of harmful language analysis}
In complement to the examples shown in section \ref{sec:LLMs} of the main body, we present additional examples examining the alignment between hate-check outputs from LLMs and VAD scores in Table \ref{tab:hate_speech}. We note here that for the hate-speech experiments, we queried the Perspective API and OpenAI Moderation API via their respective public endpoints; no local GPU inference was conducted for these tools. Whereas, the four LLMs (Mistral-8x7B, Qwen2.5-72B, GPT-OSS-120B, DeepSeek-R1-70B) were hosted locally on 4 A6000 Nvidia GPUs. No fine-tuning of any model was performed, and each run classifying all the target sentences accounted for only $\sim$1 GPU hour. 
We observe comparable trends and discrepancies in sentiment classification, particularly in how modern classifiers often fail to recognize historically situated expressions of hate or, conversely, overgeneralize from present-day connotations.


\clearpage 
\section{\CHRONOBERG Datasheet} 
\label{sec:datasheet_for_datasets}

\dssectionheader{Motivation}

\dsquestionex{For what purpose was the dataset created?}{(e.g. was there a specific task in mind? Was there a specific gap that needed to be filled? Please provide a description.)}

\dsanswer{The \CHRONOBERG dataset was created to provide a temporally structured corpus to support diachronic linguistic analysis and the evaluation of LLMs under temporal shift. While existing resources provide broad coverage, they typically lack long-term temporal structure, and are not well-suited to studying semantic drift and diachronic variation. \CHRONOBERG was designed to support tasks such as: 
\begin{itemize}
    \item Analysis of detection of discriminatory and sensitive language in historical contexts.
    \item Construction of historically grounded affective lexicons for systematic linguistic and affective analysis, and 
    \item Evaluation of temporal generalization.
\end{itemize}}

\dsquestionex{What (other) tasks could the dataset be used for?}{Are there obvious tasks for which it should not be used?}

\dsanswer{The dataset was created to provide a scalable benchmark for tasks such as:
\begin{itemize}
    \item Concept drift modelling and continual learning,
    \item Editing and unlearning to modify or update interpretations of certain English words or sentences. 
\end{itemize}}

The VAD score annotations in the dataset are NOT intended for tasks that require absolute semantic evaluation, but study relative semantic change. 

\dsquestionex{Who created this dataset (e.g., which team, research group) and on behalf of which entity ?}{(e.g., company, institution, organization)} 

\dsanswer
The creators of the dataset are: Niharika Hegde$^{1,4*}$, Subarnaduti Paul$^{2*}$,
Lars-Joel Frey$^{4}$, Manuel Brack$^{1,3\ddagger}$, Kristian Kersting$^{1,3,4,5}$, Martin Mundt$^{2\dagger}$, Patrick Schramowski$^{1,3,4,6\dagger}$.
The associated affiliations are:
$^1$German Research Center for Artificial Intelligence (DFKI), Darmstadt, Germany,
$^2$Department of Computer Science and Mathematics, University of Bremen, Bremen, Germany,
$^3$Hessian Center for AI (hessian.AI), Darmstadt, Germany,
$^4$Department of Computer Science, TU Darmstadt, Darmstadt, Germany,
$^5$Centre for Cognitive Science, TU Darmstadt, Darmstadt, Germany and
$^6$CERTAIN, Germany.

\dsquestionex{Who funded the creation of the dataset?}{If there is an associated grant, please provide the name of the grantor and the grant name and number.}

\dsanswer{No funding was received for the creation of the dataset.}

\dsquestion{Any other comments?}

\dsanswer{No further comments.}

\dssectionheader{Composition}

\dsquestionex{What do the instances that comprise the dataset represent (e.g., documents, photos, people, countries)?}{ Are there multiple types of instances (e.g., movies, users, and ratings; people and interactions between them; nodes and edges)? Please provide a description.}

\dsanswer{The instances in the \CHRONOBERG dataset represent digitized books sourced from Project Gutenberg. Each instance corresponds to: 
\begin{itemize}
    \item Full text of a book 
    \item Temporal metadata (publication year)
    \item Lexicons that capture affective sentiment dimensions such as Valence, Dominance, and Arousal.
    \item Sentence level annotations associating a VAD score. 
\end{itemize}}

\dsquestion{How many instances are there in total (of each type, if appropriate)?}

\dsanswer{\CHRONOBERG is composed of 2.7 billion tokens, representing roughly 91 million sentences from 25,061 English-language books published between 1750 and 2000, with additional metadata in the form of temporally-aligned VAD lexicons that span 335,804 words.
}

\dsquestionex{Does the dataset contain all possible instances or is it a sample (not necessarily random) of instances from a larger set?}{ If the dataset is a sample, then what is the larger set? Is the sample representative of the larger set (e.g., geographic coverage)? If so, please describe how this representativeness was validated/verified. If it is not representative of the larger set, please describe why not (e.g., to cover a more diverse range of instances, because instances were withheld or unavailable).}

\dsanswer{\CHRONOBERG is curated from Project Gutenberg, an openly accessible corpus of literary texts. Our primary focus was to derive a temporally structured dataset suitable for studying diachronic semantic drift, i.e., how word and sentence meanings evolve across centuries. To this end, we restricted our scope to Late Modern English works published between 1750 and 2000. Since Project Gutenberg's metadata on original publication date is often inaccurate or missing, we developed an inference pipeline that leverages the OpenLibrary API to obtain corrected publication years. After this process, we retained a total of 25,061 books. }

\dsquestionex{What data does each instance consist of? “Raw” data (e.g., unprocessed text or images) or features?}{In either case, please provide a description.}

\dsanswer{Each instance represents processed texts. In addition, we have also introduced temporally-aligned VAD lexicons as part of our metadata, where each instance represents processed English words.    

}

\dsquestionex{Is there a label or target associated with each instance?}{If so, please provide a description.}

\dsanswer{The dataset is grouped by year of publication. So, texts contain labels in the form of a specific year to which they belong. In addition, we have provided ``valence'', ``arousal'', and ``dominance'' annotations for each sentence in a specific time period. }

\dsquestionex{Is any information missing from individual instances?}{If so, please provide a description, explaining why this information is missing (e.g., because it was unavailable). This does not include intentionally removed information, but might include, e.g., redacted text.}

\dsanswer{There is no information missing, as we have excluded such examples from the dataset in its construction process.}

\dsquestionex{Are relationships between individual instances made explicit (e.g., users’ movie ratings, social network links)?}{If so, please describe how these relationships are made explicit.}

\dsanswer{Relationships between the individual instances are made explicit through their shared temporal alignment. Each book instance is linked to a publication year, which allows for grouping, comparison and sequential ordering across time. Instances can also be grouped into genres such as Fiction, History, Social Life, Conduct of Life, and Travel, inherited from the original Project Gutenberg corpus. At the same time, given the historical span of 1750–2000, the texts could also be categorized with respect to major historical contexts and events of the period. We did not pursue this latter categorization in depth, as we believe it requires domain expertise beyond the scope of our work. }

\dsquestionex{Are there recommended data splits (e.g., training, development/validation, testing)?}{If so, please provide a description of these splits, explaining the rationale behind them.}

\dsanswer{Yes. Although the best performing year predictor yields an uncertainty of 3-5 years, this margin is negligible at the scale of our diachronic analysis. To ensure robustness against temporal noise and to better capture semantic shifts over time, we recommend that alternately created training, validation and test sets be constructed within coarse temporal bins no smaller than 15 years.}

\dsquestionex{Are there any errors, sources of noise, or redundancies in the dataset?}{If so, please provide a description.}

\dsanswer{Yes, as mentioned above there is an uncertainty of around 3-5 years as a consequence of our publication date inference pipeline. The uncertainty originates from the fact that publication dates are often missing or refer to the time of digitization in their original online repository and alternate sources needed to be inquired. We believe the uncertainty is acceptable in light of our considered 250 year time range.}

\dsquestionex{Is the dataset self-contained, or does it link to or otherwise rely on external resources (e.g., websites, tweets, other datasets)?}{If it links to or relies on external resources, a) are there guarantees that they will exist, and remain constant, over time; b) are there official archival versions of the complete dataset (i.e., including the external resources as they existed at the time the dataset was created); c) are there any restrictions (e.g., licenses, fees) associated with any of the external resources that might apply to a future user? Please provide descriptions of all external resources and any restrictions associated with them, as well as links or other access points, as appropriate.}

\dsanswer{Yes, the dataset is self-contained. There are no access restrictions or required external resources.}

\dsquestionex{Does the dataset contain data that might be considered confidential (e.g., data that is protected by legal privilege or by doctor-patient confidentiality, data that includes the content of individuals non-public communications)?}{If so, please provide a description.}

\dsanswer{No.}

\dsquestionex{Does the dataset contain data that, if viewed directly, might be offensive, insulting, threatening, or might otherwise cause anxiety?}{If so, please describe why.}

\dsanswer{Yes. We acknowledge that the literary texts from the 1750s to the 2000s that compose \CHRONOBERG may contain sensitive content that could be offensive, insulting, or threatening to certain groups. While we do not intend to objectify anyone, we also aim to preserve the integrity of the original works without alteration and avoiding historical erasure. At the same time, this approach opens up new possibilities for exploring how emotions were directed toward specific groups during different historical periods.}

\dsquestionex{Does the dataset relate to people?}{If not, you may skip the remaining questions in this section.}

\dsanswer{Yes, we acknowledge that \CHRONOBERG contains instances of text that may refer to groups of people or individuals, either directly or indirectly. As it comprises historical literary works from the 1750s to the 2000s, some texts are curated from literal biographies of historical figures, while others relate to wars and other significant events of the period. }

\dsquestionex{Does the dataset identify any subpopulations (e.g., by age, gender)?}{If so, please describe how these subpopulations are identified and provide a description of their respective distributions within the dataset.}

\dsanswer{Yes, there are literary works that refer to or identify subpopulations by age, group or gender, considering that literary works span across different genres and include historical content.}

\dsquestionex{Is it possible to identify individuals (i.e., one or more natural persons), either directly or indirectly (i.e., in combination with other data) from the dataset?}{If so, please describe how.}

\dsanswer{Yes, in particular historical figures are explicitly identified in books containing historical non-fiction content.}

\dsquestionex{Does the dataset contain data that might be considered sensitive in any way (e.g., data that reveals racial or ethnic origins, sexual orientations, religious beliefs, political opinions or union memberships, or locations; financial or health data; biometric or genetic data; forms of government identification, such as social security numbers; criminal history)?}{If so, please provide a description.}

\dsanswer{Yes. While the dataset is composed of historical texts rather than personal records, it contains language that may be considered sensitive. This includes expressions of racial, ethnic, religious, gendered, political bias reflective of the time periods covered (1750-2000). Such content may involve discriminatory or offensive terminology, depiction of marginalized groups or outdated normative assumptions. However, the dataset itself does not contain personal identifiers, financial, health, biometric or government identification data.}

\dsquestion{Any other comments?}

\dsanswer{No further comments.}

\dssectionheader{Collection Process}

\dsquestionex{How was the data associated with each instance acquired?}{Was the data directly observable (e.g., raw text, movie ratings), reported by subjects (e.g., survey responses), or indirectly inferred/derived from other data (e.g., part-of-speech tags, model-based guesses for age or language)? If data was reported by subjects or indirectly inferred/derived from other data, was the data validated/verified? If so, please describe how.}

\dsanswer{The data was directly acquired from Project Gutenberg, an online library of copyright-free e-books from the past few centuries. It also allows easy access by mirroring their entire catalogue or downloading their e-book collection via an API. The data was directly observable as raw texts. }

\dsquestionex{What mechanisms or procedures were used to collect the data (e.g., hardware apparatus or sensor, manual human curation, software program, software API)?}{How were these mechanisms or procedures validated?}

\dsanswer{We have followed the official recommendation from Project Gutenberg to download the RDF files of the books via mirroring. The official mirror links can be found at their official website \url{https://www.gutenberg.org/help/mirroring.html}. We have also used their official repository \url{https://github.com/gutenbergtools} to interact with their resources when needed. 
}

\dsquestion{If the dataset is a sample from a larger set, what was the sampling strategy (e.g., deterministic, probabilistic with specific sampling probabilities)?}

\dsanswer{We have specifically focused on curating English texts from the period 1750–2000. However, in some cases, the original publication date of a work was either missing or inaccurately recorded. To address this, we employed an additional sampling strategy to ensure that works were properly curated and accurately categorized into their respective time intervals. }

\dsquestion{Who was involved in the data collection process (e.g., students, crowdworkers, contractors) and how were they compensated (e.g., how much were crowdworkers paid)?}

\dsanswer{Only the authors and co-authors were responsible for the collection of the data.}

\dsquestionex{Over what time-frame was the data collected? Does this time-frame match the creation time-frame of the data associated with the instances (e.g., recent crawl of old news articles)?}{If not, please describe the time-frame in which the data associated with the instances was created.}

\dsanswer{The data was curated between October 2024 and March 2025. Since the dataset consists solely of historically published literary works, its content remains unaffected by the timeline of collection and compilation. }

\dsquestionex{Were any ethical review processes conducted (e.g., by an institutional review board)?}{If so, please provide a description of these review processes, including the outcomes, as well as a link or other access point to any supporting documentation.}

\dsanswer{No}

\dsquestionex{Does the dataset relate to people?}{If not, you may skip the remaining questions in this section.}

\dsanswer{Given its historical context, the data can relate to people or groups of individuals. The literary texts constituting \CHRONOBERG represent biographies, works on important historical events between 1750 and 2000, social life, and similar aspects pertaining to historical society.}

\dsquestion{Did you collect the data from the individuals in question directly, or obtain it via third parties or other sources (e.g., websites)?}

\dsanswer{No, the data was not collected directly from any individual; rather, it was acquired from Project Gutenberg, a historical corpus of literary works, some of which may pertain to certain individuals. }

\dsquestionex{Were the individuals in question notified about the data collection?}{If so, please describe (or show with screenshots or other information) how notice was provided, and provide a link or other access point to, or otherwise reproduce, the exact language of the notification itself.}

\dsanswer{No.}

\dsquestionex{Did the individuals in question consent to the collection and use of their data?}{If so, please describe (or show with screenshots or other information) how consent was requested and provided, and provide a link or other access point to, or otherwise reproduce, the exact language to which the individuals consented.}

\dsanswer{The literary works obtained from Project Gutenberg are copyright-free and freely distributable. We have ensured that no information in these works was altered, preserving their integrity and originality. Given the historical time-frame covered by \CHRONOBERG, the individuals referenced in these texts are not available to provide consent.}

\dsquestionex{If consent was obtained, were the consenting individuals provided with a mechanism to revoke their consent in the future or for certain uses?}{If so, please provide a description, as well as a link or other access point to the mechanism (if appropriate).}

\dsanswer{Not applicable.}

\dsquestionex{Has an analysis of the potential impact of the dataset and its use on data subjects (e.g., a data protection impact analysis) been conducted?}{If so, please provide a description of this analysis, including the outcomes, as well as a link or other access point to any supporting documentation.}

\dsanswer{No such analysis with respect to data protection or privacy could be conducted as a) historical figures have long been deceased, b) information on historical figures has been disseminated in various historical works throughout time, c) Project Gutenberg has already provided a curated public archive that has excluded copy-righted and non-consensual material outside the public domain. \CHRONOBERG derives itself from Project Gutenberg and thus not induce any new impact regarding data subjects.}

\dsquestion{Any other comments?}

\dsanswer{No further comments.}

\dssectionheader{Preprocessing/cleaning/labelling}

\dsquestionex{Was any preprocessing/cleaning/labelling of the data done (e.g., discretization or bucketing, tokenization, part-of-speech tagging, SIFT feature extraction, removal of instances, processing of missing values)?}{If so, please provide a description. If not, you may skip the remainder of the questions in this section.}

\dsanswer{We have followed several steps of preprocessing and labelling of our curated raw texts from Project Gutenberg. 

\begin{itemize}
\item \textbf{Determination of original publication date:} We observed that e-books in Project Gutenberg often lack accurate publication years, or in some cases, the information is entirely missing. This step is particularly important for \CHRONOBERG, as incorrect publication years would lead to misclassification of texts into the wrong temporal intervals.  

\item \textbf{Data partitioning:} The texts are grouped by year, resulting in 250 separate splits corresponding to individual years. In addition, we created broader bins spanning 50-year intervals.  

\item \textbf{Data filtering:} We removed all non-alphanumeric characters from the texts to facilitate easier adaptation across various downstream applications.  
\end{itemize}
}

\dsquestionex{Was the “raw” data saved in addition to the preprocessed/cleaned/labelled data (e.g., to support unanticipated future uses)?}{If so, please provide a link or other access point to the “raw” data.}

\dsanswer{Yes, as a direct part of the dataset, we have made available two versions of \CHRONOBERG: one consisting of the raw texts grouped by publication year, and another with processed texts, split into annotated sentences and likewise organized by publication year.}

\dsquestionex{Is the software used to preprocess/clean/label the instances available?}{If so, please provide a link or other access point.}

\dsanswer{Yes, a public GitHub repository explaining the preprocessing, cleaning, and labelling process is available at: \url{https://github.com/paulsubarna/Chronoberg/}. Simultaneously, the dataset is available on HuggingFace: \url{https://huggingface.co/datasets/spaul25/Chronoberg}}

\dsquestion{Any other comments?}

\dsanswer{No further comments.}

\dssectionheader{Uses}

\dsquestionex{Has the dataset been used for any tasks already?}{If so, please provide a description.}

\dsanswer{The dataset has not been publicly available before. We highlight several potential downstream applications of \CHRONOBERG in our accompanying experimental analysis, including: (i) validating temporally calibrated VAD lexicons against words with documented evaluative and non-evaluative changes from literature, (ii) inspecting words and sentences within \CHRONOBERG that have undergone diachronic shifts and outline future prospects, such as (iii) unlearning or modifying specific connotations in words used in contemporary English texts.}

\dsquestionex{Is there a repository that links to any or all papers or systems that use the dataset?}{If so, please provide a link or other access point.}

\dsanswer{No.}

\dsquestion{What (other) tasks could the dataset be used for?}

\dsanswer{Beyond the experiments presented in this paper, \CHRONOBERG supports culturomics research, temporally grounded sentiment analysis and furthers research of historically situated NLP benchmarks. The VAD lexicons may also support research on social bias and stereotyping across historical periods. As part of our metadata, we also provide valence, arousal, and dominance scores for each sentence in \CHRONOBERG, aiming to capture the affective sentiment expressed in the text. However, given the sensitivity of certain sentences, we strongly caution against interpreting these scores as definitive labels of positivity or negativity.}

\dsquestionex{Is there anything about the composition of the dataset or the way it was collected and preprocessed/cleaned/labelled that might impact future uses?}{For example, is there anything that a future user might need to know to avoid uses that could result in unfair treatment of individuals or groups (e.g., stereotyping, quality of service issues) or other undesirable harms (e.g., financial harms, legal risks) If so, please provide a description. Is there anything a future user could do to mitigate these undesirable harms?}

\dsanswer{Alongside their publication year, we have also annotated the sentences in \CHRONOBERG based on their VAD scores, denoting their affective polarity. However, we acknowledge that not every negatively scored sentence based on the VAD score can be used as a harmful sentence. So, we encourage users to not treat the dataset for benchmarking hateful vs non-hateful applications. Similarly, as this data may directly or indirectly involve individuals, we discourage its use to specifically single out any individuals solely based on the VAD lexicons and affective polarity scores of the texts. }

\dsquestionex{Are there tasks for which the dataset should not be used?}{If so, please provide a description.}

\dsanswer{Please refer to the previous question.}

\dsquestion{Any other comments?}

\dsanswer{No further comments.}

\dssectionheader{Distribution}

\dsquestionex{Will the dataset be distributed to third parties outside of the entity (e.g., company, institution, organization) on behalf of which the dataset was created?}{If so, please provide a description.}

\dsanswer{Yes, the dataset is publicly available at HuggingFace \url{https://huggingface.co/datasets/spaul25/Chronoberg}, and the source code is available at GitHub \url{https://github.com/paulsubarna/Chronoberg/}}

\dsquestionex{How will the dataset will be distributed (e.g., tarball on website, API, GitHub)}{Does the dataset have a digital object identifier (DOI)?}

\dsanswer{We have made the dataset publicly available at HuggingFace: \url{https://huggingface.co/datasets/spaul25/Chronoberg} }

\dsquestion{When will the dataset be distributed?}

\dsanswer{The dataset will be distributed after de-anonymization.}

\dsquestionex{Will the dataset be distributed under a copyright or other intellectual property (IP) license, and/or under applicable terms of use (ToU)?}{If so, please describe this license and/or ToU, and provide a link or other access point to, or otherwise reproduce, any relevant licensing terms or ToU, as well as any fees associated with these restrictions.}

\dsanswer{The dataset is distributed under the BSD 2-Clause “Simplified” License. It also comes under the full list of Project Gutenberg licenses, which can be found in \url{https://www.gutenberg.org/policy/license.html} }

\dsquestionex{Have any third parties imposed IP-based or other restrictions on the data associated with the instances?}{If so, please describe these restrictions, and provide a link or other access point to, or otherwise reproduce, any relevant licensing terms, as well as any fees associated with these restrictions.}

\dsanswer{No.}

\dsquestionex{Do any export controls or other regulatory restrictions apply to the dataset or to individual instances?}{If so, please describe these restrictions, and provide a link or other access point to, or otherwise reproduce, any supporting documentation.}

\dsanswer{No.}

\dsquestion{Any other comments?}

\dsanswer{No further comments.}

\dssectionheader{Maintenance}

\dsquestion{Who will be supporting/hosting/maintaining the dataset?}

\dsanswer{The authors will be responsible for the maintenance and support of the dataset.}

\dsquestion{How can the owner/curator/manager of the dataset be contacted (e.g., email address)?}

\dsanswer{The creators of the datasets: Subarnaduti Paul and Niharika Hegde can be contacted at \href{spaul@uni-bremen.de}{spaul@uni-bremen.de} and \href{niharika.hegde@dfki.de}{niharika.hegde@dfki.de} respectively.}

\dsquestionex{Is there an erratum?}{If so, please provide a link or other access point.}

\dsanswer{There currently exists no erratum.}

\dsquestionex{Will the dataset be updated (e.g., to correct labelling errors, add new instances, delete instances)?}{If so, please describe how often, by whom, and how updates will be communicated to users (e.g., mailing list, GitHub)?}

\dsanswer{Currently, no immediate updates are envisioned. If there appears to be an urgent need to update the dataset, the authors will be responsible for uploading a new version. We will update the version at HuggingFace and communicate the news on our website and HuggingFace.}

\dsquestionex{If the dataset relates to people, are there applicable limits on the retention of the data associated with the instances (e.g., were individuals in question told that their data would be retained for a fixed period of time and then deleted)?}{If so, please describe these limits and explain how they will be enforced.}

\dsanswer{As the acquired data is copyright-free and free to distribute, we don't believe there are any applicable limits on the retention of the data associated with the instances.}

\dsquestionex{Will older versions of the dataset continue to be supported/hosted/maintained?}{If so, please describe how. If not, please describe how its obsolescence will be communicated to users.}

\dsanswer{Yes, we will continue to host and support older versions of the dataset. HuggingFace, as a platform, supports versioning.}

\dsquestionex{If others want to extend/augment/build on/contribute to the dataset, is there a mechanism for them to do so?}{If so, please provide a description. Will these contributions be validated/verified? If so, please describe how. If not, why not? Is there a process for communicating/distributing these contributions to other users? If so, please provide a description.}

\dsanswer{The code for data generation is publicly available on GitHub at \url{https://github.com/paulsubarna/Chronoberg}. Validation/Verification of future contributions will not be in the scope of the authors.
Yes, there are numerous possibilities to extend the dataset, such as:
\begin{itemize}
    \item Extending beyond the 1750–2000 time-frame, the corpus can also be viewed as an ever-growing temporal dataset of historical contexts, given copy-right laws being based on 20 and 100 year cut-offs respectively.
    \item The dataset can be extended with several other languages besides Late Modern English. 
\end{itemize}   
}

\dsquestion{Any other comments?}

\dsanswer{No further comments.}

\end{document}